\pdfoutput=1

\documentclass[11pt]{article}

\usepackage[]{ACL2023}

\usepackage{times}
\usepackage{latexsym}

\usepackage[T1]{fontenc}

\usepackage[utf8]{inputenc}

\usepackage{microtype}
\usepackage{amsfonts}
\usepackage{xcolor}
\usepackage{amsmath}
\usepackage{siunitx}
\usepackage{dcolumn}
\newcolumntype{d}[1]{D{.}{.}{#1}}
\usepackage{arydshln}
\usepackage{multirow}
\usepackage{booktabs}
\usepackage{graphicx}
\usepackage{algorithm, algpseudocode}
\usepackage{graphicx}
\usepackage{verbatim}
\usepackage{stmaryrd}
\usepackage{trimclip}
\usepackage{subcaption}
\usepackage{url}
\usepackage{xspace}
\usepackage[normalem]{ulem}
\useunder{\uline}{\ul}{}

\usepackage[most]{tcolorbox}
\definecolor{block-gray}{gray}{0.85}

\newtcolorbox{myquote}{colback=block-gray,grow to right by=-5mm,grow to left by=-5mm,boxrule=0pt,boxsep=0pt,breakable}

\usepackage{inconsolata}

\newcommand{\X}{\texttt{[X]}\xspace}
\newcommand{\gi}{\texttt{[$g_i$]}\xspace}
\newcommand{\aj}{\texttt{[$a_j$]}\xspace}
\newcommand{\ck}{\texttt{[$c_k$]}\xspace}

\newcommand{\vGEP}{$\overrightarrow{\mathrm{GEP}}$\xspace}

\title{Auditing Gender Presentation Differences in Text-to-Image Models}

\author{Yanzhe Zhang, Lu Jiang$^{1,2}$, Greg Turk, Diyi Yang$^3$ \\
Georgia Institute of Technology, $^1$Google Research \\ $^2$Carnegie Mellon University, $^3$Stanford University \\
  \texttt{z\_yanzhe@gatech.edu, turk@cc.gatech.edu}\\ \texttt{$^2$lujiang@cmu.edu, $^3$diyiy@cs.stanford.edu }}

\begin{document}
\maketitle
\begin{abstract}
Text-to-image models, which can generate high-quality images based on textual input, have recently enabled various content-creation tools. Despite significantly affecting a wide range of downstream applications, the distributions of these generated images are still not fully understood, especially when it comes to the potential stereotypical attributes of different genders. 
In this work, we propose a paradigm (Gender Presentation Differences) that utilizes fine-grained self-presentation attributes to study how gender is presented differently in text-to-image models.
By probing gender indicators in the input text (e.g., ``\emph{a woman}'' or ``\emph{a man}''), we quantify the frequency differences of presentation-centric attributes (e.g., ``\emph{a shirt}'' and ``\emph{a dress}'') through human annotation and introduce a novel metric: GEP.\footnote{\textbf{GEP}: \textbf{GE}nder \textbf{P}resentation Differences. This study uses this term to refer specifically to the attribute-level presentation differences between images generated from different gender indicators. Note that the definition of GEP is not built on the common usage of gender presentation (gender expression, used to distinguish from gender identity. \citealt{wikipedia_2022}).}
Furthermore, we propose an automatic method to estimate such differences. The automatic GEP metric based on our approach yields a higher correlation with human annotations than that based on existing CLIP scores, consistently across three state-of-the-art text-to-image models.
Finally, we demonstrate the generalization ability of our metrics in the context of gender stereotypes related to occupations. \footnote{Project website: \url{https://salt-nlp.github.io/GEP/}.}
\end{abstract}

\section{Introduction}

Being able to generate photorealistic images and artwork, text-to-image models have achieved remarkable progress recently \citep{ramesh2022hierarchical, saharia2022photorealistic, chang2023muse}, enabling many downstream applications such as image editing \citep{kawar2022imagic, hertz2022prompttoprompt, brooks2022instructpix2pix}, inpainting \citep{lugmayr2022repaint, Rombach_2022_CVPR}, and style transfer \citep{Rombach_2022_CVPR}.
Although this defines a promising pattern for commercial content creation, the lack of understanding and evaluation of the generated images hinders the deployment of text-to-image models in real-world scenarios due to potential biases and stereotypes embedded in the text-to-image models.
For example, an early version of DALLE-2 often generated men with light skin colors for the input of ``\emph{A photo of a CEO.}'' \citep{openai_2022}. Such stereotypes are amplified in text-to-image models compared to real-world distributions \citep{bianchi2022easily}. However, how different social groups are represented unequally in text-to-image models is far from comprehensively studied. 

Taking gender as an example, most prior studies prompt the model with ``\emph{A photo of \X.}'' where ``\emph{\X}'' refers to occupations (``\emph{a CEO}'') or descriptors (``\emph{an attractive person}'', ``\emph{a person with a beer}'') \citep{cho2022dalleval, bansal2022texttoimage, bianchi2022easily, fraser2023a}. 
These approaches automatically classify all generated images into gender categories and measure bias using the relative gender frequencies. 
However, a person's gender should not be determined nor predicted solely by appearance \citep{american2015guidelines, zimman2019trans}, which makes these methods fundamentally inappropriate for the task.
Additionally, existing classification systems perform poorly for transgender individuals \citep{scheuerman2019computers}.
To mitigate such issues, our work studies this problem from a different perspective:
\begin{quote}
    \textit{When probing different genders in the text input, how will text-to-image models alter the person's presence in the generated image?}
\end{quote}

By the person's presence, we refer to the presence of presentation-related attributes, such as whether the person is wearing ``\emph{a shirt}'' or ``\emph{a dress}''.
In this way, we avoid appearance-based gender classification, which is subjective and raises ethical concerns. Instead, we examine concrete and objective \textbf{attribute-wise differences} between images generated by text-to-image models with different gender-specific prompts.
Note that, \emph{we aim to provide a neutral description of attribute differences present in these generated images, and suggest such differences as an objective lens for practitioners to use to understand potential issues exhibited by text-to-image models, without any presuppositions of genders in these images}. 
Different from prior works, we prompt the model with ``\emph{A man/woman \X.}'' where ``\emph{\X}'' refers to predefined contexts (e.g., ``\emph{sitting at a table}'', ``\emph{riding a bike}'') and manually annotate the frequency of attributes we are interested in the generated images. \footnote{Similar to prior works \citep{schumann2021step, cho2022dalleval, bansal2022texttoimage}, we only consider binary gender while constructing our text prompts. We acknowledge this as a  limitation and further discuss it in the Ethics section.}
We define such frequency differences between genders as gender presentation differences, for which we introduce a new quantitative metric, GEP. Specifically, we first build the GEP vectors, each dimension representing the difference in one attribute. We then calculate its normalized $\ell_1$ norm as the GEP score to summarize the magnitude of such differences for different models.
The GEP vector helps determine which attributes differ more between genders, while the GEP score indicates which model demonstrates more gender differences overall.

Extensive evaluation using human annotation for all these text-to-image models is impractical, as collecting (or reusing) ground truth images for every attribute of interest \citep{bianchi2022easily} is not scalable. In prior works, state-of-the-art text-image matching models like CLIP \citep{radford2021learning} are used for zero-shot gender classification to quantify gender biases. However, such approaches often struggle with more nuanced features like skin colors \citep{cho2022dalleval, bansal2022texttoimage}. In fact, we found that 
simply using CLIP to calculate cross-modal similarity performs poorly in detecting attributes, when estimating the GEP vector and GEP score.
To this end, we propose to 1) train attribute classifiers on the shared space of CLIP using text captions only and 2) use such classifiers to classify the CLIP embedding of generated images. These cross-modal classifiers maintain the CLIP baseline's scalable property while substantially improving the correlation with human annotations.

To summarize, our contributions are:
1) we formulate gender presentation differences and introduce the GEP metric based on a set of representative presentation attributes and contexts.
2) we analyze the proposed two metrics on three state-of-the-art open-access (or API-access) text-to-image models based on human annotations.
3) we develop a more reliable automatic estimation of the GEP vector and GEP score than prior approaches.
4) we show that GEP can also reveal attribute-wise gender stereotypes related to occupations, while prior work mainly categorized gender.

\begin{figure*}[t]
\centering
\resizebox{1.88\columnwidth}{!}{\includegraphics{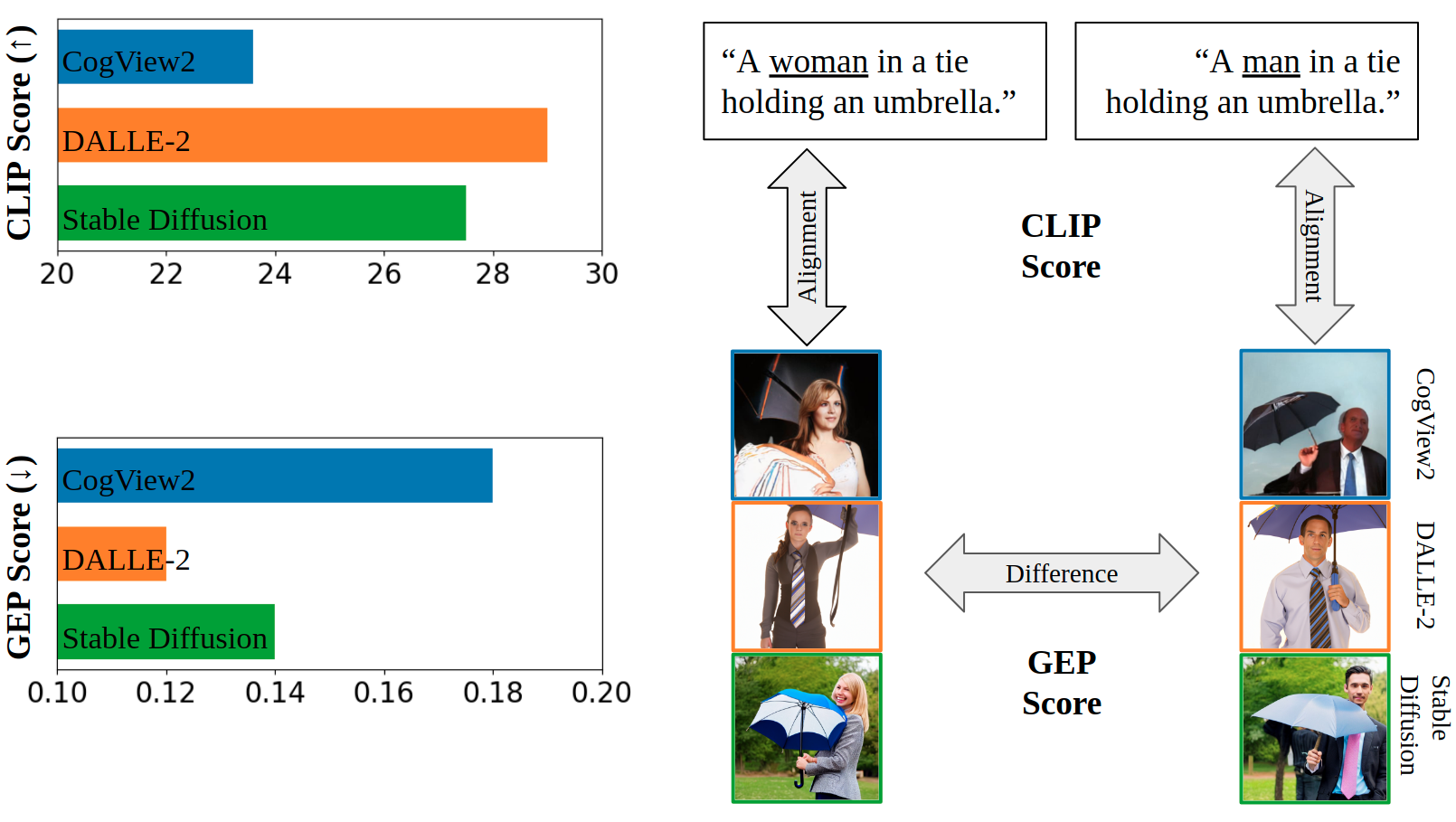}}
\caption{\label{fig-sum} Based on our prompts, we report the CLIP score and the proposed GEP score of three state-of-the-art text-to-image models. The GEP score offers a different perspective to evaluate the text-to-image models compared to the existing CLIP score, as it considers the attribute-wise differences caused by specifying different genders. Specifically, we find that DALLE-2 (API) presents gender more similarly than other models.}
\end{figure*}

\section{Related Work}

\paragraph{Text-to-image models}
Similar to language modeling, where the output space is a relatively small discrete space, one line of text-to-image research tokenizes an image into a sequence of discrete vectors \citep{van2017neural}. Based on such vectorization, CogView2 \citep{ding2022cogview2} first generates low-res images through autoregressive generation, then applies two super-resolution modules to output high-res images. Parti \citep{yu2022scaling} further shows that scaling up autoregressive models can generate photorealistic images with world knowledge.
Another line of text-to-image research, built on denoising diffusion models \citep{ho2020denoising}, has made great strides in the recent past.
DALLE-2 \citep{ramesh2022hierarchical} builds text-to-image models based on CLIP \citep{radford2021learning} embeddings followed by upsampling models. While running the denoising process in the pixel space needs large memory, Stable Diffusion \citep{rombach2021highresolution} proposes to run the denoising process in latent space, which enables high-quality image generation on small GPUs. Imagen \citep{saharia2022photorealistic} further finds that generic large language models like T5 \citep{2020t5} are effective for text-to-image generation. 
However, some of these cutting-edge text-to-image models are not open source, leaving researchers with only a few relatively weak public versions \citep{kakaobrain2021minDALL-E} to work with \citep{cho2022dalleval, bansal2022texttoimage}. With the recent advance of Stable Diffusion and the API access of DALLE-2, we build our analysis on the top of state-of-the-art public-accessible text-to-image models.

\paragraph{Bias in Text-to-image models} As discussed in \citet{saharia2022photorealistic} and \citet{openai_2022}, there are some social biases and stereotypes in the generated images of these models, which is still underexplored now. \citet{cho2022dalleval} probe the text-to-image models with occupations and human-related objects and analyze the generated people's genders and skin colors. \citet{bianchi2022easily} further observe that the stereotypes related to professions in generated images are amplified compared to the real-world distributions of occupations. For example, although women nurses are in the majority in real life,\footnote{About 90\% of registered nurses self-identified as women, according to \url{https://www.bls.gov/cps/cpsaat11.htm}.}\footnote{We are using ``women nurses'' instead of ``female nurses'', as ``female'' has biological overtones \citep{norris_2019}. More discussions can be found at \url{https://www.merriam-webster.com/words-at-play/lady-woman-female-usage}.} nearly all nurses generated by the model are women, which is extremely imbalanced.
Beyond that, \citet{Struppek2022TheBA} show that text-to-image models are subject to cultural biases, which can be easily triggered by replacing input characters with non-Latin characters.
\citet{bansal2022texttoimage} find that adding ethical intervention to the text input effectively reduces these biases.

In contrast to this prior work, we study fine-grained presentation differences. We frame this as ``difference'' rather than ``bias'' because we do not assume that all attributes should be equally distributed across gender. However, presentation differences in text-to-image models need to be studied systematically. Otherwise, they can reinforce existing presentation stereotypes, such as ``men in suits'' and ``women in dresses'', or even introduce new stereotypes as the generated content becomes part of the culture \citep{martin2014spontaneous}.

\paragraph{Evaluation of Text-to-image models}
The most common metric to evaluate text-to-image models is Fréchet Inception Distance \citep{NIPS2017_8a1d6947}, which needs a large enough number of ground truth images to calculate the distribution and is not suitable for fine-grained evaluation. Beyond that, human annotations are the gold standard for evaluation, where annotators are asked to label the image quality \citep{saharia2022photorealistic}, the relation between objects \citep{conwell2022testing}, genders, skin colors, culture \citep{cho2022dalleval, bansal2022texttoimage} and other subtle features specified by the text prompt \citep{Leivada2022DALLE2F, rassin2022dalle}. For automatic evaluations, CLIP \citep{radford2021learning} is the preferred model to measure the matching between texts and images \citep{park2021benchmark, hessel2021clipscore}. 
State-of-the-art object detectors are also used to benchmark the model's understanding of object counts and spatial relationships \citep{cho2022dalleval, Gokhale2022BenchmarkingSR}.
Following prior works \citep{cho2022dalleval, bansal2022texttoimage}, we use the CLIP similarity score (with calibration) as our baseline. Beyond that, we propose to train cross-modal classifiers based on the CLIP output space shared by texts and images, which only needs texts as training data but can better distinguish attributes in images. 

\begin{table*}[t]
\small
\centering
\begin{tabular}{ll}
\toprule \toprule
\textbf{Query to ConceptNet}     & \textbf{Items in ConceptNet (Attributes in $A$)}                                                                                            \\ \midrule
IsA footwear            & boots (``\emph{in boots}''), slippers (``\emph{in slippers}'')                                                             \\ \midrule
IsA trousers/dress      & \begin{tabular}[c]{@{}l@{}} jeans (``\emph{in jeans}''), shorts (``\emph{in shorts}''), slacks (``\emph{in slacks}''), dress (``\emph{in a dress}''), \\ skirt (``\emph{in a skirt}'') \end{tabular} \\ \midrule
IsA clothes/attire/coat & suit (``\emph{in a suit}''), shirt (``\emph{in a shirt}''), uniform (``\emph{in uniform}''), jacket (``\emph{in a jacket}'')                   \\ \midrule
IsA accessory           & hat (``\emph{in a hat}''), tie (``\emph{with a tie}''), mask (``\emph{with a mask}''), gloves (``\emph{with gloves}'')    \\ \bottomrule                    
\end{tabular}
\caption{\label{table-attribute} The attribute set $A$. Attributes are categorized by the query used to build the attribute set. Query ``IsA trousers/dress'' retrieves items that are either trousers or dresses.}
\end{table*}

\begin{table}[t]
\centering
\begin{tabular}{c}
\toprule \toprule
\textbf{Contexts in $C$}        \\ \midrule
``\emph{sitting at a table}''       \\
``\emph{sitting on a bed}''         \\
``\emph{standing on a skateboard}'' \\
``\emph{standing next to a rack}''  \\
``\emph{riding a bike}''            \\
``\emph{riding a horse}''           \\
``\emph{laying on the snow}''       \\
``\emph{laying on a couch}''        \\
``\emph{walking through a forest}'' \\
``\emph{walking down a sidewalk}''  \\
``\emph{holding up a smartphone}''  \\
``\emph{holding an umbrella}''      \\
``\emph{jumping into the air}''     \\
``\emph{jumping over a box}''       \\
``\emph{running across the park}''  \\
``\emph{running on the beach}''     \\
\bottomrule  
\end{tabular}
\caption{\label{table-context} The context set $C$.}
\end{table}

\section{Problem Definition}

\paragraph{Genders, Attributes, and Contexts}
To assess gender presentation differences, we use three elements to construct our text prompts: a set of gender indicators $G = \{g_1, \cdots, g_m \}$, a set of human-related attributes $A = \{a_1, \cdots, a_n \}$, and a set of contexts $C = \{c_1, \cdots, c_p \}$.
The definition of $G$, $A$, and $C$ is scalable, flexible, and can be adapted for different use cases. In this work, without loss of generality, we define $G$, $A$, and $C$ as follows:
\begin{itemize}
    \item $G$: We use $G = \{$ ``\emph{A woman}'', ``\emph{A man}''$\}$ as our gender indicators, which are the most widely used gender indicators. Note that $G$ can include non-binary gender indicators or occupations used in prior works, while we focus on women and men since these two genders are the most common in the text-to-image training corpus.
    \item $A$ (Table \ref{table-attribute}): We query ConceptNet \citep{speer2016conceptnet} with different types of clothing (e.g., `\emph{footwear}', `\emph{trousers}') and the relation ``IsA'' to get a list for each type of clothing. After sorting each list using the word frequency \citep{robyn_speer_2022_7199437} of each item, we apply manual filtering to get the final set of \textbf{15 attributes} fulfilling three constraints: 1) visible, 2) context-agnostic, and 3) common.
    \item $C$ (Table \ref{table-context}): We randomly retrieve human-related captions from the COCO dataset \citep{lin2014microsoft}, then get the short contexts (typically \emph{verb + preposition + noun}) for each caption. We apply manual filtering to get the final set of \textbf{16 contexts}  by ensuring the final set contains 1) diverse actions, 2) various objects, and 3) different scenarios.
\end{itemize}
$G$ defines the group between which we study the differences, and $A$ specifies the attributes on which we evaluate the differences. Beyond these, we increase the complexity of text-to-image generation by adding contexts in $C$ to the text prompts.

\paragraph{Gender Presentation Differences}
Based on $G$, $A$, and $C$, we consider \textbf{two types} of gender presentation differences:
\begin{itemize}
    \item \textbf{Neutral:} One does not specify any attributes in the input. We prompt the text-to-image model with ``\gi \ck'' (e.g., ``\emph{A man sitting at a table.}'') to get an image set $S(g_i, c_k)$. Then for each generated image, we check the existence of \textbf{every attribute} in $A$. Grouping the results by genders, for each gender $g_i$, we get a vector $ [f_{i1}, \cdots, f_{in} ]$ where $f_{ij}$ denotes the frequency of attribute $a_j$ appears in the images generated from gender $g_i$.
    \item \textbf{Explicit:} Specify attributes in the input. We prompt the text-to-image model with ``\gi \aj \ck'' (e.g., ``\emph{A man in boots standing on a skateboard.}'') to get an image set $S(g_i, a_j, c_k)$. Then for each generated image, we check whether the generated images contain the \textbf{corresponding attribute} in the input. Grouping the results by genders, for each gender $g_i$, we have a vector $ [f_{i1}, \cdots, f_{in} ]$ where $f_{ij}$ denotes the frequency of attribute $a_j$ appears in the images generated from the combination of gender $g_i$ and attribute $a_j$.
\end{itemize}

Specifically, for gender $g_i$ and attribute $a_j$, $f_{ij}$ is calculated by
\begin{align}
\label{eqa-0}
    \begin{split}
        f_{ij} = \frac{\sum_{I \in S} \mathbf{existence}(I, a_j)}{|S|}
    \end{split}
\end{align}
where $S=\bigcup_{k=1}^{p} S(g_i, c_k)$ for the neutral setting, or $S=\bigcup_{k=1}^{p} S(g_i, a_j, c_k)$ for the explicit setting. $\mathbf{existence}(I, a_j)$ returns $1$ if attribute $a_j$ exists in $I$, otherwise returns $0$. The values for attribute frequencies $f_{ij}$ are thus in the range $0$ to $1$. Here we use human annotation as the $\mathbf{existence}$ function, but later we will seek to replace human annotations with automatic measures.

The two settings target different gender presentation differences: the \emph{neutral} setting reveals the attribute difference between naturally presented genders without specifications. In contrast, the \emph{explicit} setting shows the difference in associating different genders with each attribute.

Given one setting of a model, to reflect the fine-grained difference between women ($g_1$) and men ($g_2$) on all attributes $\{a_1, \cdots, a_n \}$, we define the GEP metric as:
\begin{align}
\label{eqa-1}
    \begin{split}
        \overrightarrow{\mathrm{GEP}} & = [v_1, \cdots, v_n], \\
        \mathrm{GEP} & = \frac{1}{n} \| \overrightarrow{\mathrm{GEP}}\|_1
    \end{split}
\end{align}
where $v_j = f_{1j} - f_{2j}$ denotes the attribute-wise difference.
The GEP vector (\vGEP) helps us to analyze and compare the presentation differences in various attributes. By normalizing the $\ell_1$ norm of \vGEP, the GEP score enables comparison of the overall presentation differences between different models and settings, where a lower score indicates the two genders are presented more similarly.

In the following sections, we use $\overrightarrow{\mathrm{GEP}}\mathrm{\textsubscript{human}}$ and GEP\textsubscript{human} to denote the GEP vector and GEP score based on the human annotation of frequencies, and use $\overrightarrow{\mathrm{GEP}}\mathrm{\textsubscript{auto}}$ and GEP\textsubscript{auto} in general terms to refer to automatic estimation of GEP vectors and scores.

\begin{figure*}[ht]
\centering
\resizebox{2.0\columnwidth}{!}{\includegraphics{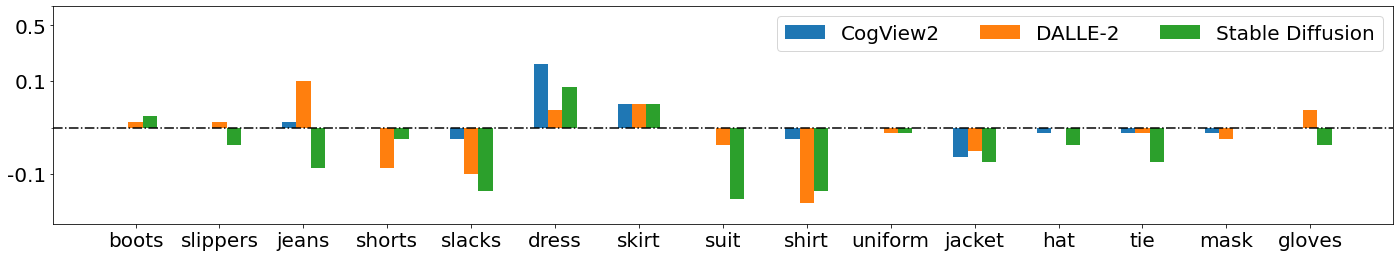}}
\resizebox{2.0\columnwidth}{!}{\includegraphics{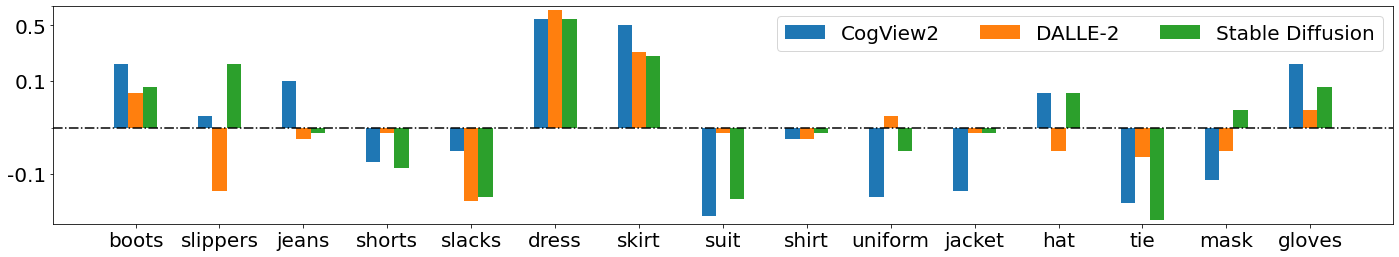}}
\caption{\label{fig-two-human} The human-annotated GEP vectors $\overrightarrow{\mathrm{GEP}}\mathrm{\textsubscript{human}}$ for three models in the neutral setting (up) and the explicit setting (bottom). The y axes are presentation differences in symmetric log scaling.}
\end{figure*}

\section{Automatic Estimation of GEP}
The goal of proposing the GEP vector/score is to quantify the gender presentation differences when we compare the text-to-image models. However, carrying out large-scale human annotations whenever new models are trained is apparently not practical. 
Given a group of images $I_1, \cdots, I_z$, we simplify Equation \ref{eqa-0} and rewrite frequency $f_a$ of attribute $a$ as (for the simplicity of notation, we ignore the gender, context, and setting):
\begin{align}
\label{eqa-3}
    \begin{split}
        f_a = \frac{1}{z}\sum_{l=1}^z \mathbf{existence}(I_l, a)
    \end{split}
\end{align}
where $\mathbf{existence}(I_l, a)$ returns $1$ if annotators find attribute $a$ in $I_l$, otherwise returns $0$.
$f_a$ is then utilized to calculate $\overrightarrow{\mathrm{GEP}}\mathrm{\textsubscript{human}}$ and GEP\textsubscript{human}.

This section explores how to estimate $f_a$ to automatically calculate $\overrightarrow{\mathrm{GEP}}$ and GEP. The estimation is not expected to return exactly $\overrightarrow{\mathrm{GEP}}\mathrm{\textsubscript{human}}$ and GEP\textsubscript{human} but should correlate well with the differences annotated by humans.
For notational convenience, we use $C\mbox{-}f_a$, $CC\mbox{-}f_a$, $CLS\mbox{-}f_a$ to denote different estimation approaches, based on which we can automatically build GEP vectors $\overrightarrow{\mathrm{GEP}}\mathrm{\textsubscript{C}}$, $\overrightarrow{\mathrm{GEP}}\mathrm{\textsubscript{CC}}$, $\overrightarrow{\mathrm{GEP}}\mathrm{\textsubscript{CLS}}$ and GEP scores GEP\textsubscript{C}, GEP\textsubscript{CC}, GEP\textsubscript{CLS} without human annotations.
Note that for all $\overrightarrow{\mathrm{GEP}}\mathrm{\textsubscript{auto}}$ and GEP\textsubscript{auto}, the scale is different from $\overrightarrow{\mathrm{GEP}}\mathrm{\textsubscript{human}}$ and GEP\textsubscript{human}, since we use CLIP similarity and probabilities predicted by classifiers to replace human-annotated frequency. Though the scale is different, a good automatic estimation should give the same or similar ranking in the comparison between attributes and between models (e.g., Fig \ref{fig-scatter-stable}). 

\subsection{CLIP similarity} 
Most prior works \citep{cho2022dalleval, bansal2022texttoimage} achieve zero-shot classification on the top of text-image matching model CLIP \citep{radford2021learning}.
Similarly, a straightforward way to reflect $f_a$ is to use the CLIP similarity score to replace the $\mathbf{existence}$ function:
\begin{align}
    \begin{split}
        C\mbox{-}f_a = \frac{1}{z}\sum_{l=1}^z cos(\mathbf{C}(I_l), \mathbf{C}(a)) 
    \end{split}
\end{align}
where $\mathbf{C}$ denotes the CLIP model that can take image input $I_l$ or textual input $a$. For instance, $a$ can be ``\emph{a dress}'', ``\emph{a suit}'', ignoring the prepositions in attributes in Table \ref{table-attribute} to represent the attribute of interest better.
We choose not to add the prefix like ``A photo of'' \citep{hessel2021clipscore, cho2022dalleval, bansal2022texttoimage}, since image generation is not constrained by this prefix.

\paragraph{Calibration} As prior works classify gender by comparing $cos(\mathbf{C}(I_l), \mathbf{C}($``female''$))$ and $cos(\mathbf{C}(I_l), \mathbf{C}($``male''$))$, it is not obvious how to adapt such comparisons to our scenario since it is hard to frame the non-existence of attributes into descriptions. For example, the contrast between ``\emph{not a dress}'' and ``\emph{a dress}'' is ineffective. Alternatively, we try to ensure the attribute $a$ is more similar to the images compared to irrelevant texts. Specifically, we use a reference string $r(a)$ to calibrate the CLIP similarity:
\begin{align}
    \begin{split}
        CC\mbox{-}f_a = & \frac{1}{z} [ \sum_{l=1}^z cos(\mathbf{C}(I_l), \mathbf{C}(a)) \\
         & - cos(\mathbf{C}(I_l), \mathbf{C}(r(a)))]
    \end{split}
\end{align}
where $cos(\mathbf{C}(I_l), \mathbf{C}(r(a)))$ can be interpreted as a dynamic threshold for $a$. By default, we set $r(a)$ to ``\emph{an object}'' \citep{bansal2022texttoimage} for all attributes, which is found to be empirically effective. We provide a discussion on this in the evaluation section.

\subsection{Our Approach: Cross-Modal Classifiers}
As discussed, one trivial way to detect all attributes is to collect enough images of each attribute to train classifiers, which is not scalable. However, though collecting high-quality images containing certain attributes is hard, creating sentences containing certain attributes (words) is easy. Note that trained on contrastive loss, the representations of images and texts are well aligned (not perfectly aligned as discussed in \citet{ramesh2022hierarchical}) in the output space of CLIP, implying that we do not need the real images to obtain the image embeddings \citep{Nukrai2022TextOnlyTF, Gu2022ICB}. This enables us to train classifiers in such a shared space using text embeddings first, then use trained classifiers to classify image embeddings.

To build a classifier for attribute $a$, we build a positive set $P$ of sentences containing attribute $a$ and a negative set $N$ of sentences without attribute $a$. Precisely, we follow the pattern of ``\texttt{[$\hat{g}$]}\xspace \texttt{[$a$]}\xspace \texttt{[$\hat{c}$]}\xspace'' to create the positive set and the pattern of ``\texttt{[$\hat{g}$]}\xspace \texttt{[$\hat{c}$]}\xspace'' to create the negative set. We use $\hat{g} \in \hat{G} = \{$ ``\emph{A man}'', ``\emph{A woman}'', ``\emph{A person}''$\}$, and all $\hat{c} \in \hat{C} = C$, which is the same context set used in image generation.

Then we train a simple logistic regression model $\mathbf{cls}$ on top of CLIP embeddings using the binary cross entropy loss:
\begin{align}
    \begin{split}
        L = \sum_{(x,y) \in P \cup N} (  -y \log (\mathbf{cls}(\mathbf{C}(x))) \\ 
        - (1 - y) \log (1 - \mathbf{cls}(\mathbf{C}(x))))
    \end{split}
\end{align}
where $y = 1$ for positive examples, $y = 0$ for negative examples.
Thus, $\mathbf{cls}(\mathbf{C}())$ can output high probabilities for sentences containing attribute $a$. Furthermore, since the output space is aligned for images and texts, we assume it can also assign high probabilities for images that contain attribute $a$, which allows us to create a metric as follows:
\begin{align}
    \begin{split}
        CLS\mbox{-}f_a = \frac{1}{z}\sum_{l=1}^z \mathbf{cls}(\mathbf{C}(I_l)))
    \end{split}
\end{align}
In practice, we separately train ten classifiers using different random seeds and average their predictions for $\mathbf{cls}(\mathbf{C}(I_l))$. This ensemble won't cost much extra time since logistic regression models are super fast to train.

Unlike the CLIP similarity, which aggregates the information in all dimensions of CLIP embeddings to one scalar, such a classifier-based approach can distinguish the dimension-wise differences between embeddings to determine the existence of attribute $a$ better.

\begin{table}[t]
\centering
\begingroup
\setlength{\tabcolsep}{4.5pt} 
\renewcommand{\arraystretch}{1.0} 
\begin{tabular}{lcccc} \toprule \toprule
\textbf{}        & \multicolumn{2}{c}{Neutral} & \multicolumn{2}{c}{Explicit} \\
\textbf{}        & GEP ($\downarrow$)          & CS ($\uparrow$)           & GEP ($\downarrow$)           & CS ($\uparrow$)           \\ \midrule
\textbf{CogView} & 0.02         & 23.5         & 0.18          & 23.6         \\
\textbf{DALLE}   & 0.05         & 26.3         & 0.12          & 29.0         \\
\textbf{Stable}  & 0.07         & 26.5         & 0.14          & 27.5         \\ \bottomrule
\end{tabular}
\endgroup
\caption{\label{table-gep-score-human} GEP\textsubscript{human} and CLIPScore (CS). To save space, we use \textbf{CogView} for Cogview2, \textbf{DALLE} for DALLE-2, and \textbf{Stable} for Stable Diffusion. Note that images generated by the DALL-E API may be subject to internal filter algorithms.}
\end{table}

\begin{figure*}[t]
\centering
\resizebox{2.0\columnwidth}{!}{\includegraphics{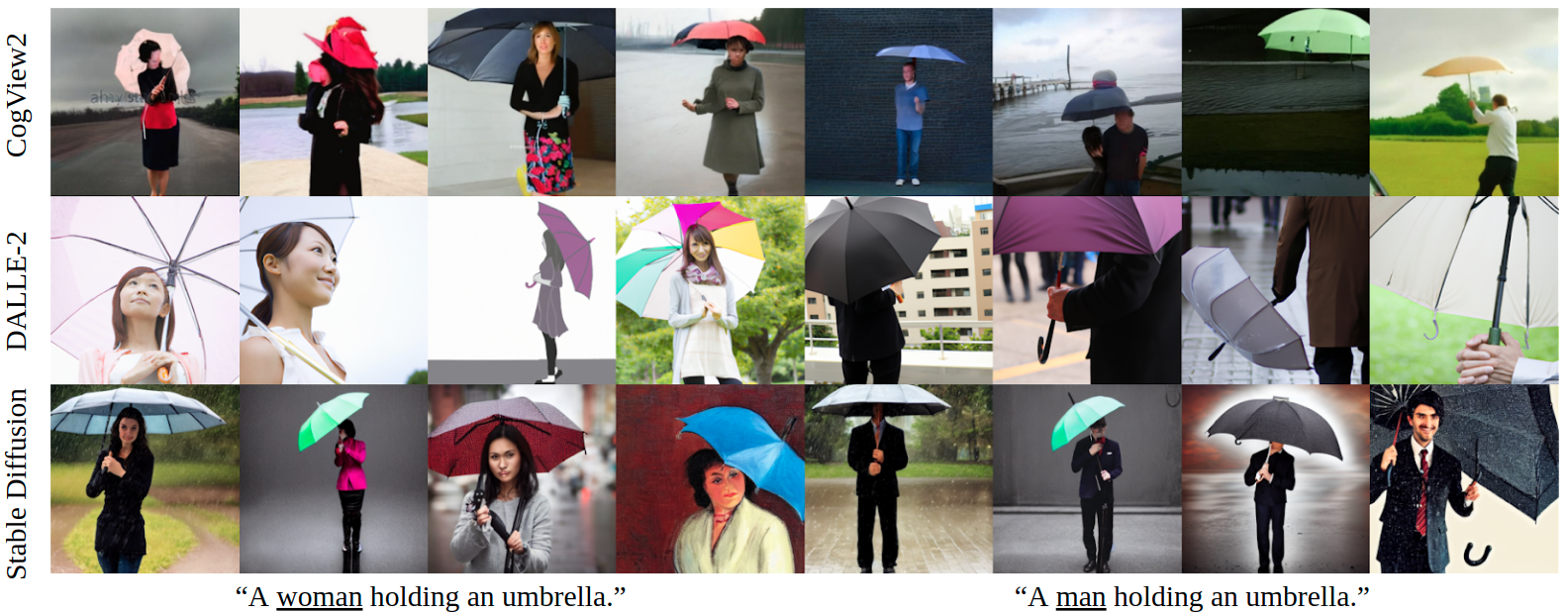}}
\caption{\label{fig-example-base-umbrealla} Examples of gender presentation differences in the neutral setting. Images are generated from ``\emph{A woman/man holding an umbrella.}''. In generated images, women are more likely to wear dresses and shirts, while men are more likely to wear suits.}
\end{figure*}
\begin{figure*}[t]
\centering
\resizebox{2.0\columnwidth}{!}{\includegraphics{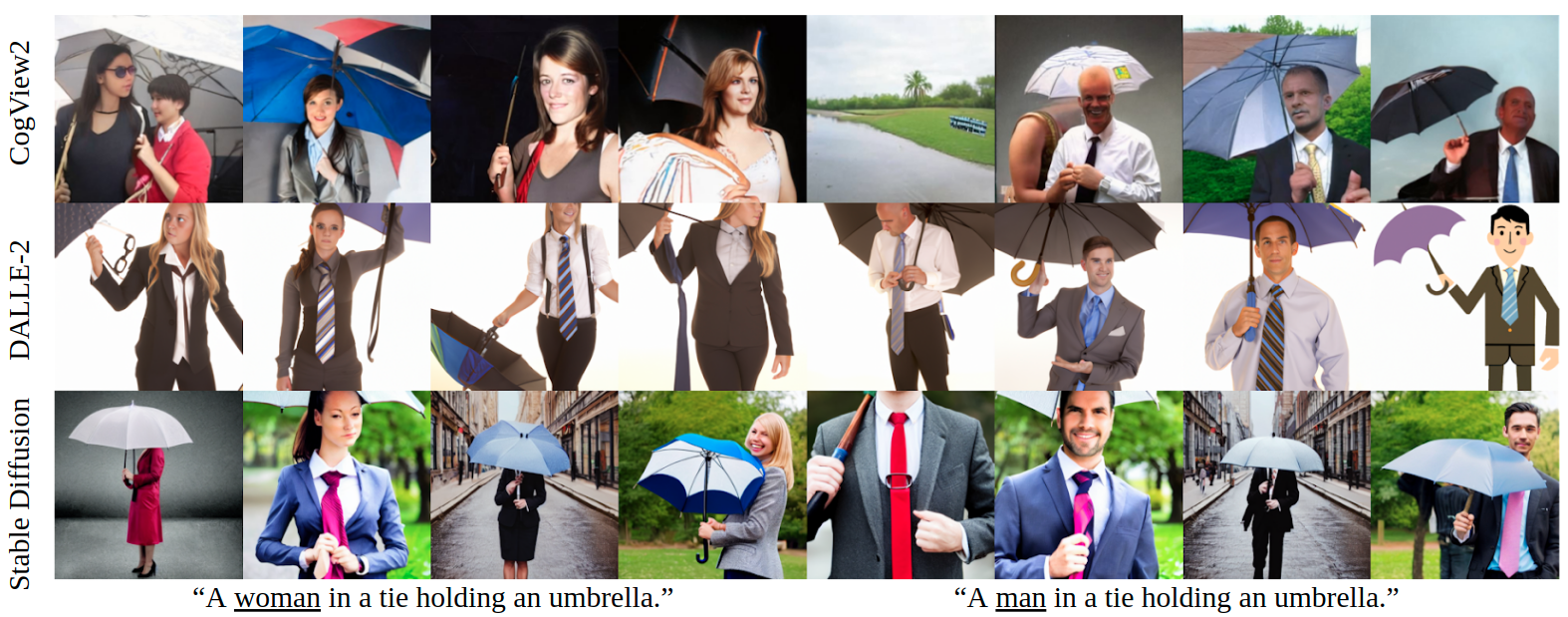}}
\caption{\label{fig-example-comb-tie} Examples of gender presentation differences in the explicit setting. Images are generated from ``\emph{A woman/man in a tie holding an umbrella.}''. The gender presentation differences still exist even we mention the attribute in the input, as CogView2 and Stable Diffusion struggle to generate women in a tie.}
\end{figure*}

\section{Analysis of GEP\textsubscript{human}}

\paragraph{Text-to-Image Models} We select three state-of-the-art text-to-image models for evaluation: 1) \textbf{CogView2} \citep{ding2022cogview2}: an open-sourced vectorization-based text-to-image model trained on 30 million text-image pairs. 2) \textbf{DALLE-2} \citep{ramesh2022hierarchical}: State-of-the-art diffusion-based text-to-image models with API access. Note that images generated by this API might be deliberately filtered or diversified, thus diverging from the original output distribution. 3) \textbf{Stable Diffusion} \citep{rombach2021highresolution}: The most popular open-sourced text-to-image model with latent diffusion, which is trained on LAION-5B \citep{schuhmann2022laion5b} with aesthetic score thresholds. For detailed model configuration, see Appendix \ref{sec:appendix-config}.

\paragraph{Image Generation}
According to the previous section, we have 32 prompts for the neutral setting and 480 for the explicit setting. For each prompt, we generate five images (e.g., $|S(g_i, c_k)|=5$). For each model, we generate 160 images for the neutral setting and 2400 for the explicit setting.

\paragraph{Human Annotation}
We annotate all 15 attributes for each image in the neutral setting. For each image in the explicit setting, we only annotate the attribute used to generate that image. We complete the annotation on Amazon Mechanism Turk.\footnote{https://www.mturk.com/} For a pair of $($image $I,$ attribute $a)$ that need annotation, we ask three workers, ``\emph{Do you see a person [a]?}'' (e.g., ``\emph{Do you see a person in a dress?}''). The majority of the three annotations determine the ground truth labels. Krippendorff's alpha between annotators is 0.81. See Appendix \ref{sec:appendix-mturk} for details on workers and annotations.

\paragraph{Results}
We plot $\overrightarrow{\mathrm{GEP}}\mathrm{\textsubscript{human}}$ in Figure \ref{fig-two-human}, and GEP\textsubscript{human} in Table \ref{table-gep-score-human}. We also report the CLIPScore, which evaluates the alignment between the text prompts and the generated images in Table \ref{table-gep-score-human} as references. Please refer to the appendix for detailed frequencies (Table \ref{table-base} and \ref{table-combination}) and frequency differences (Table \ref{table-base-diff} and \ref{table-combination-diff}).

In the neutral setting, the most significant differences are on ``\emph{jeans}'', ``\emph{shorts}'', ``\emph{slacks}'', ``\emph{dress}'', ``\emph{suit}'', and ``\emph{shirt}''.
We show examples in Figure \ref{fig-example-base-umbrealla}, \ref{fig-example-base-bike} and \ref{fig-example-base-couch}.
Some presentation differences coincide across models and with common stereotypes: ``\emph{dress}'' and ``\emph{skirt}'' have higher frequencies in women than men, and ``\emph{suit}'', ``\emph{shirt}'', and ``\emph{slacks}'' have higher frequencies with men than women. \footnote{Note that the word ``women'' here refers to images generated using ``a woman'' in the prompt. Same for the word ``men''. We do not make assumptions about the genders of the people generated.}
However, the magnitude of differences varies considerably from model to model. For example, there are greater differences related to ``\emph{suit}'' and ``\emph{tie}'' in the images generated by Stable Diffusion than DALLE-2, while the difference associated with ``\emph{shirt}'' is smaller. More than that, attribute ``\emph{jeans}'' shows opposite tendencies in different models: it is generated more frequently with women using DALLE-2 while more frequently with men using Stable Diffusion. We hypothesize differences in training corpora can lead to such opposite tendencies. Though CogView2 has the lowest GEP\textsubscript{human} (2\%), the conclusion that CogView2 presents genders the most equally is not comprehensive: CogView2 has the lowest CLIPScore (23.5) in this setting, indicating relatively weak text-image alignment and low image quality for both genders. More concretely, we find that fewer attributes are detected in images generated by CogView2 (average frequency 6\%) compared to other models (average frequency $\geq$ 9\%), making the smallest difference reasonable.

In the explicit setting, the magnitudes of presentation differences are \textbf{amplified} for most attributes compared to the neutral setting: $15/15$ attributes for CogView2, $9/15$ attributes for DALLE-2, $11/15$ attributes for Stable Diffusion, probably due to attributes being generated more frequently while mentioning them, which results in more considerable differences between genders. For example, it's easier for text-to-image models to associate ``\emph{boots}'' and ``\emph{gloves}'' with women than men (Figure \ref{fig-example-comb-boots}), which is not revealed in the neutral setting. The largest difference comes from ``\emph{dress}'', where all three models failed to associate it with men (Figure \ref{fig-example-comb-dress}). Similarly, models struggle to associate ``\emph{tie}'' with women (Figure \ref{fig-example-comb-tie}). For those cases where differences are not amplified, such as ``\emph{jeans}'', ``\emph{shorts}'', and ``\emph{suit}'' generated by DALLE-2, we find that they are almost perfectly generated in both genders (frequencies $\geq$ 97\%). With the strongest ability to associate given attributes with both genders (average frequency $\geq$ 88\%), DALLE-2 demonstrates the lowest GEP\textsubscript{human} (12\%) among all three models, while it also demonstrates the best image quality with the highest CLIPScore (29.0).\footnote{We discuss its limitation in the Limitation section.}

\begin{table*}[t]
\centering
\begingroup
\setlength{\tabcolsep}{4.5pt} 
\renewcommand{\arraystretch}{1.0} 
\begin{tabular}{lcccccc}
\toprule \toprule
\multicolumn{1}{c}{} & \multicolumn{3}{c}{\textbf{Neutral}}            & \multicolumn{3}{c}{\textbf{Explicit}}                    \\ \midrule
\multicolumn{1}{c}{} & \textbf{CogView} (15) & \textbf{DALLE} (15) & \textbf{Stable} (15) & \textbf{CogView} (15) & \textbf{DALLE} (15) & \textbf{Stable} (15) \\
$\overrightarrow{\mathrm{\textbf{GEP}}}\mathrm{\textsubscript{C}}$                 & 0.416/\textbf{0.408}            & 0.413/0.491          & 0.348/0.262           & 0.714/0.464   & \textbf{0.309}/\textbf{0.289}          & 0.567/0.189           \\
$\overrightarrow{\mathrm{\textbf{GEP}}}\mathrm{\textsubscript{CC}}$                & 0.416/0.080            & 0.413/0.378          & 0.348/0.262           & 0.676/0.134             & 0.329/0.200          & 0.490/0.491  \\
$\overrightarrow{\mathrm{\textbf{GEP}}}\mathrm{\textsubscript{CLS}}$        & \textbf{0.499}/0.167   & \textbf{0.567}/\textbf{0.600} & \textbf{0.638}/\textbf{0.318}  & \textbf{0.733}/\textbf{0.607}            & 0.232/0.189 & \textbf{0.702}/\textbf{0.600}  \\ \bottomrule
\end{tabular}
\endgroup
\caption{\label{table-crr-base-combination} Correlation between the automatic GEP vectors $\overrightarrow{\mathrm{GEP}}\mathrm{\textsubscript{C}}$, $\overrightarrow{\mathrm{GEP}}\mathrm{\textsubscript{CC}}$, $\overrightarrow{\mathrm{GEP}}\mathrm{\textsubscript{CLS}}$ and $\overrightarrow{\mathrm{GEP}}\mathrm{\textsubscript{human}}$. For each $\overrightarrow{\mathrm{GEP}}\mathrm{\textsubscript{auto}}$ on each model, we report Kendall's Tau ($\uparrow$) / MCC ($\uparrow$) and highlight the strongest correlation. The numbers in parenthesis are the number of examples to calculate the correlation. We show the number of examples used to calculate the correlation in parentheses.}
\end{table*}

\begin{table}[t]
\centering
\small
\begingroup
\setlength{\tabcolsep}{4pt} 
\renewcommand{\arraystretch}{1.2} 
\begin{tabular}{lccc}
\toprule \toprule
                    & \multicolumn{3}{c}{\textbf{Artificial}}  \\ \midrule
\multicolumn{1}{c}{} & \textbf{CogView} (450) & \textbf{DALLE} (600) & \textbf{Stable} (600) \\
$\overrightarrow{\mathrm{\textbf{GEP}}}\mathrm{\textsubscript{C}}$           & 0.526/0.487         & 0.709/0.642         & 0.665/0.604 \\
$\overrightarrow{\mathrm{\textbf{GEP}}}\mathrm{\textsubscript{CC}}$         & 0.549/0.428         & 0.730/0.630         & 0.704/0.653  \\
$\overrightarrow{\mathrm{\textbf{GEP}}}\mathrm{\textsubscript{CLS}}$ & \textbf{0.590}/\textbf{0.520} & \textbf{0.780}/\textbf{0.693} & \textbf{0.748}/\textbf{0.725}  \\ \bottomrule
\end{tabular}
\endgroup
\caption{\label{table-crr-artificial} Correlation between the automatic GEP vectors $\overrightarrow{\mathrm{GEP}}\mathrm{\textsubscript{C}}$, $\overrightarrow{\mathrm{GEP}}\mathrm{\textsubscript{CC}}$, $\overrightarrow{\mathrm{GEP}}\mathrm{\textsubscript{CLS}}$ and $\overrightarrow{\mathrm{GEP}}\mathrm{\textsubscript{human}}$ on the artificial datasets. For each $\overrightarrow{\mathrm{GEP}}\mathrm{\textsubscript{auto}}$ on each model, we report  Kendall's Tau ($\uparrow$) / MCC ($\uparrow$). The number of examples used to calculate the correlation is shown in parentheses after each model name.}
\end{table}

\begin{table*}[t]
\centering
\small
\begingroup
\setlength{\tabcolsep}{2pt} 
\renewcommand{\arraystretch}{1.0} 
\begin{tabular}{lccccccc}
\toprule \toprule
          & \multicolumn{3}{c}{\textbf{Neutral}}                   & \multicolumn{3}{c}{\textbf{Explicit}}            & \multicolumn{1}{c}{\textbf{Tau} ($\uparrow$)} \\ \midrule
\textbf{} & \textbf{CogView} & \textbf{DALLE} & \textbf{Stable} & \textbf{CogView} & \textbf{DALLE} & \textbf{Stable} & \textbf{}                        \\ 
\textbf{GEP}\textsubscript{C}        & 5.15$\times 10^{-3}$ (\#3)             & 5.06$\times 10^{-3}$ (\#2)           & 4.80$\times 10^{-3}$ (\#1)             & 9.81$\times 10^{-3}$ (\#6)            & 9.76$\times 10^{-3}$ (\#5)           & 7.08$\times 10^{-3}$ (\#4)            & 0.466                            \\
\textbf{GEP}\textsubscript{CC}           & 4.49$\times 10^{-3}$ (\#1)             & 5.68$\times 10^{-3}$ (\#3)           & 4.77$\times 10^{-3}$ (\#2)            & 1.10$\times 10^{-2}$ (\#6)            & 1.04$\times 10^{-2}$ (\#5)          & 6.58$\times 10^{-3}$ (\#4)            & 0.733                            \\
\textbf{GEP}\textsubscript{CLS}            & 3.79$\times 10^{-2}$ (\#1)             & 4.26$\times 10^{-2}$ (\#2)           & 4.47$\times 10^{-2}$ (\#3)            & 9.25$\times 10^{-2}$ (\#6)             & 5.18$\times 10^{-2}$ (\#4)           & 6.32$\times 10^{-2}$ (\#5)            & \textbf{1.000}                            \\ \midrule
\textbf{GEP}\textsubscript{human}             & 0.02 (\#1)             & 0.05 (\#2)           & 0.07 (\#3)            & 0.18 (\#6)             & 0.12 (\#4)           & 0.14 (\#5)            &                                  \\ \bottomrule                          
\end{tabular}
\endgroup
\caption{\label{table-avgabs-pred} The automatic GEP scores GEP\textsubscript{C}, GEP\textsubscript{CC}, GEP\textsubscript{CLS} and GEP\textsubscript{human} on both settings of three models. We report Kendall's Tau correlation between GEP\textsubscript{auto} and GEP\textsubscript{human}. As discussed, the scale of GEP\textsubscript{auto} is different, what we care about is whether they can predict the rank (in the parentheses) of GEP\textsubscript{human}.}
\end{table*}

\begin{table}[t]
\centering
\begin{tabular}{lccc}
\toprule \toprule
\textbf{}               & \textbf{CogView} & \textbf{DALLE} & \textbf{Stable} \\ \midrule
$C\mbox{-}f_a$           & 0.812            & 0.923          & 0.860           \\
$CC\mbox{-}f_a$        & 0.842            & 0.930          & 0.881           \\
$CLS\mbox{-}f_a$ & \textbf{0.865}   & \textbf{0.950} & \textbf{0.892}  \\ \bottomrule
\end{tabular}
\caption{\label{table-avg-rocauc} Averaged Area under ROC curve (AUC) for $C\mbox{-}f_a$, $CC\mbox{-}f_a$, and $CLS\mbox{-}f_a$. We report the detailed ROC-AUC ($\uparrow$) for each attribute in Table \ref{table-rocauc-detail}.}
\end{table}

\begin{table}[t]
\small
\centering
\begingroup
\setlength{\tabcolsep}{5pt} 
\renewcommand{\arraystretch}{1.2} 
\begin{tabular}{lccc}
\toprule \toprule
$\overrightarrow{\mathrm{\textbf{GEP}}}\mathrm{\textsubscript{CC}}$        & \textbf{CogView} & \textbf{DALLE} & \textbf{Stable} \\ \midrule
``\emph{an object}'' & 0.549/0.428   & 0.730/0.630  & 0.704/0.653  \\
``\emph{}''         & 0.517/0.410    & 0.723/0.639  & 0.691/0.632           \\
``\emph{clothes}''   & 0.475/0.436   & 0.661/0.575  & 0.652/0.476           \\ \midrule
$\overrightarrow{\mathrm{\textbf{GEP}}}\mathrm{\textsubscript{C}}$          & 0.526/0.487            & 0.709/0.642          & 0.665/0.604           \\ \midrule
\end{tabular}
\endgroup
\caption{\label{table-reference} The effect of using different reference strings for $\overrightarrow{\mathrm{GEP}}\mathrm{\textsubscript{CC}}$. We report Kendall's Tau ($\uparrow$) / MCC ($\uparrow$) on three artificial datasets. 
}
\end{table}

\begin{table}[t]
\small
\begingroup
\setlength{\tabcolsep}{4pt} 
\renewcommand{\arraystretch}{1.2} 
\begin{tabular}{lccc}
\toprule \toprule
Ablation               & \textbf{CogView} & \textbf{DALLE} & \textbf{Stable} \\ \midrule
No Ensemble            & 0.574/0.501 & 0.763/0.729 & 0.709/0.689 \\
$N=N_{random}$                 & 0.544/0.326 & 0.584/0.535 & 0.667/0.618 \\
$\hat{C}=C_{random}$           & 0.602/0.506 & 0.787/0.712 & 0.761/0.690 \\
$\hat{G}=G$                    & 0.605/0.524 & 0.768/0.626 & 0.763/0.672 \\ \midrule
$\overrightarrow{\mathrm{GEP}}\mathrm{\textsubscript{CLS}}$              & 0.590/0.520 & 0.780/0.693 & 0.748/0.725 \\ \bottomrule
\end{tabular}
\endgroup
\caption{\label{table-ablation} Ablation Study of $\overrightarrow{\mathrm{GEP}}\mathrm{\textsubscript{CLS}}$. We report Kendall's Tau ($\uparrow$) / MCC ($\uparrow$) on three artificial datasets.}
\end{table}

\section{Evaluation of GEP\textsubscript{auto}}

\paragraph{Metrics}
For automatic GEP vectors and scores, we evaluate whether they can approximate $\overrightarrow{\mathrm{GEP}}\mathrm{\textsubscript{human}}$ and GEP\textsubscript{human} to ease the comparison of gender presentation differences, i.e., showing a high correlation with $\overrightarrow{\mathrm{GEP}}\mathrm{\textsubscript{human}}$ and GEP\textsubscript{human}.
1) Given two GEP vectors $\overrightarrow{\mathrm{GEP}}\mathrm{\textsubscript{human}}$ and $\overrightarrow{\mathrm{GEP}}\mathrm{\textsubscript{auto}}$, we evaluate the automatic one by Kendall rank correlation coefficient (Kendall's tau, \citealp[]{kendall1938new}) and Matthews correlation coefficient (MCC, \citealp[]{matthews1975comparison}).\footnote{MCC only takes binary variables as input. So we convert values in both GEP vectors to binary values using zero as the threshold to calculate MCC.} The former evaluates whether $\overrightarrow{\mathrm{GEP}}\mathrm{\textsubscript{auto}}$ gives the same ordering of differences as $\overrightarrow{\mathrm{GEP}}\mathrm{\textsubscript{human}}$ so we can use it to compare the differences between attributes. The latter evaluates whether $\overrightarrow{\mathrm{GEP}}\mathrm{\textsubscript{auto}}$ gives the same sign (positive or negative) of differences as $\overrightarrow{\mathrm{GEP}}\mathrm{\textsubscript{human}}$ on each attribute so that we can judge which gender is preferred. We provide extra discussions about the metric evaluation in the Appendix. 2) To evaluate each GEP\textsubscript{auto}, we calculate its estimation on both settings of all three models, which is then used to calculate the Kendall rank correlation with GEP\textsubscript{human}. 
Comparing Kendall's tau informs us which estimation approach works best.

\paragraph{Data} 
For $\overrightarrow{\mathrm{GEP}}\mathrm{\textsubscript{auto}}$: 1) Real-world Examples: Based on annotated differences in Figure \ref{fig-two-human}, we first evaluate $\overrightarrow{\mathrm{GEP}}\mathrm{\textsubscript{auto}}$ within each setting of each model. The correlation between the 15 numbers in $\overrightarrow{\mathrm{GEP}}\mathrm{\textsubscript{auto}}$ and the 15 numbers in $\overrightarrow{\mathrm{GEP}}\mathrm{\textsubscript{human}}$ is calculated in each case.
2) Artificial Examples: The number of real-world examples used to calculate the correlation is relatively few. So we manually create an artificial dataset for each model, which consists of various magnitudes of frequency differences for each attribute. In particular, given an attribute, we randomly sample a number between -1 and 1 as ``an artificial difference'', create two groups (one for women, one for men) of generated images that have the corresponding frequency difference, and finally, use such two groups of images as ``an example of difference''. We build relatively large-scale artificial datasets of presentation differences by running the same process multiple times. We create 450 examples for CogView2, 600 for DALLE-2, and 600 for Stable Diffusion. See Appendix \ref{sec:appendix-dataset} for details on creating artificial datasets. For each model, one example counts as one dimension in $\overrightarrow{\mathrm{GEP}}\mathrm{\textsubscript{human}}$ for artificial datasets so that we can evaluate $\overrightarrow{\mathrm{GEP}}\mathrm{\textsubscript{auto}}$ robustly.
For GEP\textsubscript{auto}, we use the six ground truth GEP\textsubscript{human} in Table \ref{table-gep-score-human}. \smallskip

\noindent
\textbf{Experiment Details}
For the CLIP model, we use \texttt{ViT-L/14}. For the logistic regression model used in cross-modal classifiers, we use SGD, with learning rate $1e-3$, maximum iterations $5000$, validation fraction $10\%$, and early stopping with $5$ iterations in scikit-learn library \citep{pedregosa2012scikitlearn}. The training set for each attribute consists of 96 examples (48 positive examples, 48 negative examples). After extracting CLIP features for images and texts, evaluating all 16 attributes through the ensemble takes about 1.6 seconds on CPUs. We use the Stable Diffusion artificial datasets to select all hyperparameters as a validation set.

\begin{figure*}[ht]
\centering
\resizebox{2.0\columnwidth}{!}{\includegraphics{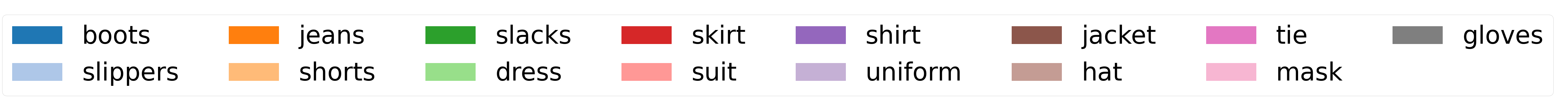}}
\begin{subfigure}{.5\textwidth}
  \centering
  \includegraphics[width=1\linewidth]{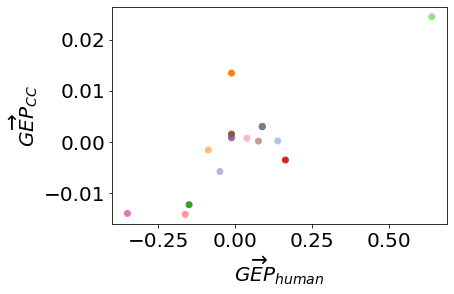}
  \caption{CLIP-Calibrated}
  \label{fig:sub1}
\end{subfigure}%
\begin{subfigure}{.5\textwidth}
  \centering
  \includegraphics[width=0.98\linewidth]{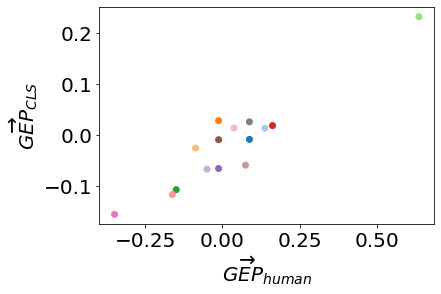}
  \caption{Classifiers}
  \label{fig:sub2}
\end{subfigure}
\caption{\label{fig-scatter-stable} (a) The correlation between the $\overrightarrow{\mathrm{GEP}}\mathrm{\textsubscript{human}}$ and $\overrightarrow{\mathrm{GEP}}\mathrm{\textsubscript{CC}}$. (b) The correlation between the $\overrightarrow{\mathrm{GEP}}\mathrm{\textsubscript{human}}$ and $\overrightarrow{\mathrm{GEP}}\mathrm{\textsubscript{CLS}}$. Both are in the explicit setting of Stable Diffusion.}
\end{figure*}

\begin{figure*}[t]
\centering
\resizebox{2.0\columnwidth}{!}{\includegraphics{legend.png}}
\begin{subfigure}{.5\textwidth}
  \centering
  \includegraphics[width=1\linewidth]{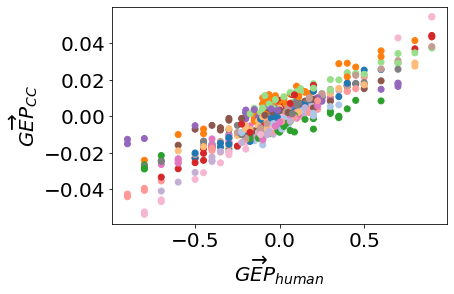}
  \caption{CLIP-Calibrated}
  \label{fig:sub3}
\end{subfigure}%
\begin{subfigure}{.5\textwidth}
  \centering
  \includegraphics[width=1\linewidth]{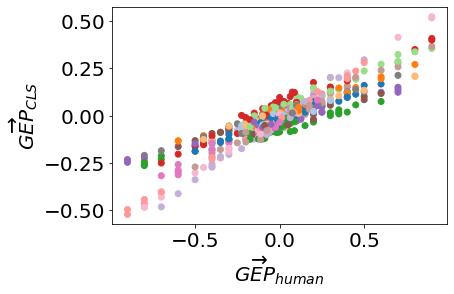}
  \caption{Classifiers}
  \label{fig:sub4}
\end{subfigure}
\caption{\label{fig-scatter-stable-artificial} The correlation between the $\overrightarrow{\mathrm{GEP}}\mathrm{\textsubscript{human}}$ and the differences predicted by CLIP similarity with calibration $\overrightarrow{\mathrm{GEP}}\mathrm{\textsubscript{CC}}$ and our cross-modal classifiers $\overrightarrow{\mathrm{GEP}}\mathrm{\textsubscript{CLS}}$ on the artificial datasets of Stable Diffusion.}
\end{figure*}

\subsection{Results}

\paragraph{$\overrightarrow{\mathrm{\textbf{GEP}}}\mathrm{\textsubscript{auto}}$ Evaluation}
We show the correlation results on the real-world examples in Table \ref{table-crr-base-combination}. Compared to $\overrightarrow{\mathrm{GEP}}\mathrm{\textsubscript{C}}$ and $\overrightarrow{\mathrm{GEP}}\mathrm{\textsubscript{CC}}$, $\overrightarrow{\mathrm{GEP}}\mathrm{\textsubscript{CLS}}$ strongly correlates with human annotations in most cases, especially when the gender presentation differences are considerable in the explicit setting. We plot the correlation between $\overrightarrow{\mathrm{GEP}}\mathrm{\textsubscript{human}}$ and $\overrightarrow{\mathrm{GEP}}\mathrm{\textsubscript{CC}}$, $\overrightarrow{\mathrm{GEP}}\mathrm{\textsubscript{CLS}}$ in Figure \ref{fig-scatter-stable} to demonstrate the improvement. There are several cases where all automatic metrics failed to show strong correlations, such as the DALLE-2 model in the explicit setting. Taking a closer look at $\overrightarrow{\mathrm{GEP}}\mathrm{\textsubscript{human}}$ in Table \ref{table-base-diff} and \ref{table-combination-diff}, we hypothesize the failure is due to the presentation differences of many attributes (e.g., ``\emph{shorts}'', ``\emph{suit}'', and ``\emph{shirt}'') being very close to 0. For example, minor differences like -0.01 and -0.02 are challenging for automatic metrics to predict the rankings and the signs correctly.
Moreover, comparing Figure \ref{fig-scatter-stable-artificial} to Figure \ref{fig-scatter-stable}, artificial datasets allow for more comprehensive testing of correlations with more data points.
Showing the correlation results on our artificial datasets in Table \ref{table-crr-artificial}, we find that adding calibration improves $\overrightarrow{\mathrm{GEP}}\mathrm{\textsubscript{C}}$ in evaluating Stable Diffusion, while it cannot improve the MCC on DALLE-2 and CogView2.
However, $\overrightarrow{\mathrm{GEP}}\mathrm{\textsubscript{CLS}}$ consistently and significantly strengthens both metrics on all models, demonstrating its generalization ability.

\paragraph{GEP\textsubscript{auto} Evaluation}
The evaluation of $\overrightarrow{\mathrm{GEP}}\mathrm{\textsubscript{auto}}$ is an attribute-level evaluation, which evaluates whether the automatic estimation can help us compare the differences between attributes and predict the sign of each attribute's presentation difference.
Differently, the evaluation of GEP\textsubscript{auto} is a model-level evaluation, which evaluates whether the automatic estimation help with comparing the magnitude of gender presentation differences produced by each model.
We compare GEP\textsubscript{auto} in Table \ref{table-avgabs-pred} and report Kendall's tau.
Consistent with $\overrightarrow{\mathrm{GEP}}\mathrm{\textsubscript{auto}}$ evaluation, we find that GEP\textsubscript{CLS} perfectly correlates with the GEP\textsubscript{human} in terms of Kendall's tau (1.00), outperforming GEP\textsubscript{C} (0.47) and GEP\textsubscript{CC} (0.73) baselines.
On the one hand, CLIP similarity depends on factors other than attributes, such as image quality, so it might not have the generalization ability to compare results across models. 
On the other hand, our cross-modal classifiers demonstrate better consistency across models.

\subsection{Analysis}

\paragraph{How good are $C\mbox{-}f_a$, $CC\mbox{-}f_a$, and $CLS\mbox{-}f_a$?} 
In the previous discussions, we mainly discuss the correlation between automatic GEP and human-annotated GEP. However, we haven't discussed how well $C\mbox{-}f_a$, $CC\mbox{-}f_a$, and $CLS\mbox{-}f_a$ represent the ground truth existence of attributes (Equation \ref{eqa-3}). For images generated by each model, 320 images (from two settings) are labeled with the presence or absence of each attribute. Use those annotations as references and automatic metrics like $CLS\mbox{-}f_{a}$ as predictions, we report the ROC-AUC score of each attribute for each model in Appendix Table \ref{table-rocauc-detail}, and the averaged scores over attributes in Table \ref{table-avg-rocauc}. A higher ROC-AUC score suggests that the existence and absence of attributes separate the predicted values well, which means the predicted values are better indicators for $f_a$. On average, the ROC-AUC scores are high, suggesting they are reliable indicators of the existence of attributes. While adding calibration slightly increases the CLIP similarity score baseline, using cross-modal classifiers brings the strongest performance, consistent with the results in Table \ref{table-crr-artificial}.
Furthermore, in the Appendix, we provide some grounded examples for $CLS\mbox{-}f_a$ to demonstrate its correctness.

\paragraph{The design of reference strings $r(a)$} Here, we compare results using different reference strings in the calibration. Besides the ``\emph{an object}'' we use by default, we consider an empty string and ``\emph{clothes}'': the former could be seen as another irrelevant reference string, while the latter is related to our attributes. We show the results in Table \ref{table-reference} and find that the choice of reference strings is important for calibration. Even using an empty string helps improve the performance compared to no calibration, while using ``\emph{clothes}'' largely hurts the performance. This finding confirms our hypothesis that an irrelevant string works better as a reference string, with which contrasting the similarity is more effective.

\paragraph{Ablation Study on $\overrightarrow{\mathrm{\textbf{GEP}}}\mathrm{\textsubscript{CLS}}$} We study which parts of the design in cross-modal classifiers lead to the superior performance of $\overrightarrow{\mathrm{GEP}}\mathrm{\textsubscript{CLS}}$ using artificial datasets. 1) By default, we train ten classifiers for each attribute and average their predictions as an ensemble to calculate $CLS\mbox{-}f_a$. We remove the ensemble and find a consistent drop in performance across models and metrics. We assume training multiple classifiers can discover more complementary useful features which produce a more reliable attribute detection through the ensemble. 2) Instead of following the pattern of ``\texttt{[$\hat{g}$]}\xspace \texttt{[$\hat{c}$]}\xspace'' to create the negative sets for training, we use random captions from the COCO dataset to perform an ablation study. Performance drops significantly as a result of this change. Thus, context alignment between the positive and negative sets is critical to train classifiers that can distinguish attributes. 3) Originally, we use the context set $C$, which is used to generate images, to construct the sentences for training. To test the contribution of contexts, we use $\hat{C}=C_{random}$, randomly sampled contexts from the COCO dataset, to construct aligned positive and negative sets. We find no noticeable performance change, either increase or decrease. Thus, our classifiers do not rely on training from image-related contexts, suggesting good generalization ability. 4) We use  $\hat{G} = G = \{$ ``\emph{A woman}'', ``\emph{A man}''$\}$ to construct the training sentences as an ablation study. By comparing with the performance of using $\hat{G} = \{$ ``\emph{A woman}'', ``\emph{A man}'',``\emph{A person}''$\}$, we find adding the word ``\emph{A person}'' essentially helps with correctly predicting the sign of differences (higher MCC on DALLE-2 and Stable Diffusion), which we assume a gender-neutral term helps the classifier to be more gender-neutral.

\begin{figure}[t]
\centering
\resizebox{1.0\columnwidth}{!}{\includegraphics{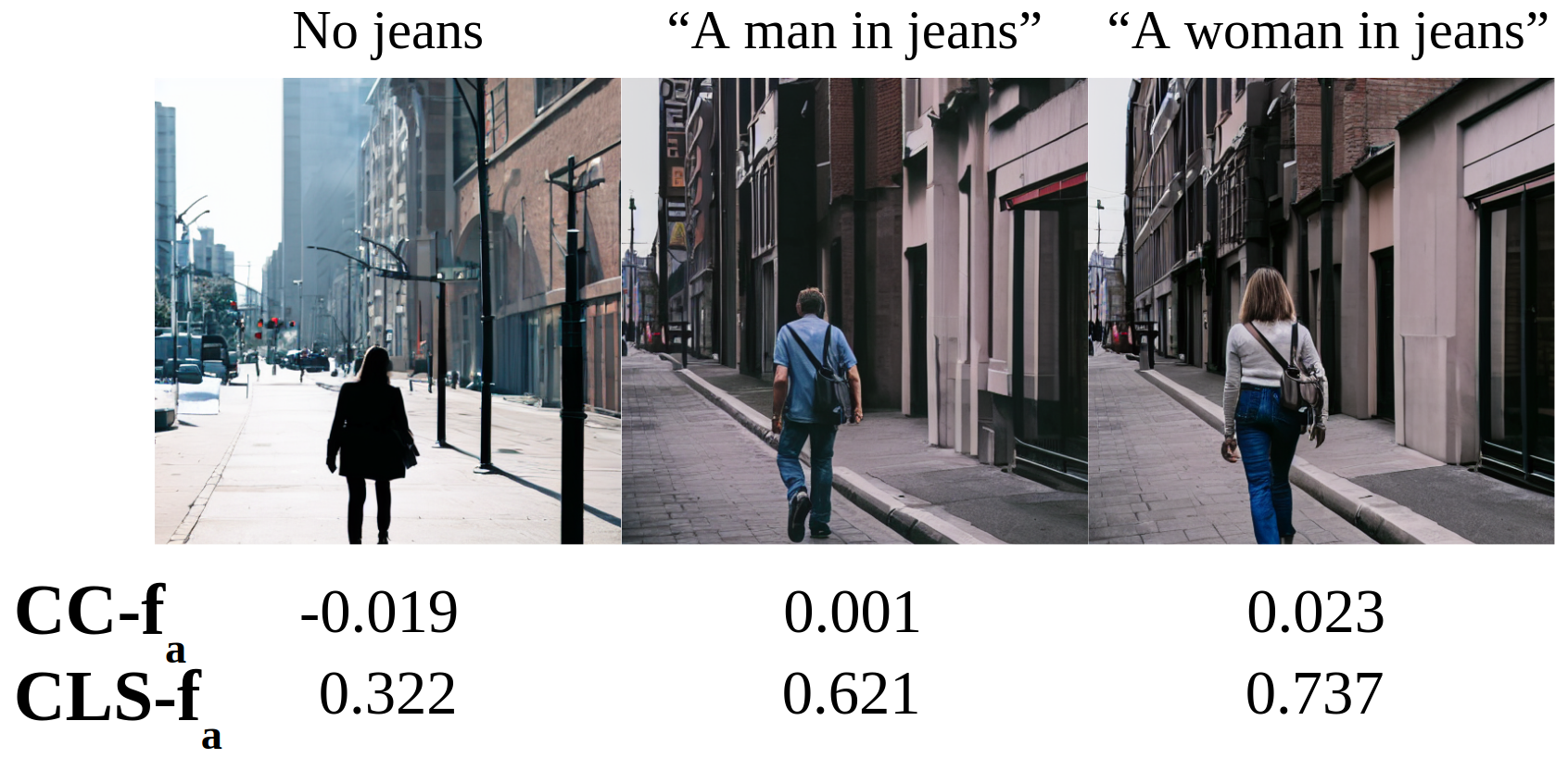}}
\caption{\label{fig-improvement} Examples for the improvement analysis. The relative difference between genders is larger using $CC\mbox{-}f_a$ than $CLS\mbox{-}f_a$, making it harder to compare the existence of different attributes in $\overrightarrow{\mathrm{GEP}}\mathrm{\textsubscript{CC}}$.
}
\end{figure}

\paragraph{Improvement Analysis of $\overrightarrow{\mathrm{\textbf{GEP}}}\mathrm{\textsubscript{CLS}}$}
By comparing the two figures in Figure \ref{fig-scatter-stable}, which depicts the explicit setting of Stable Diffusion, we analyze the performance improvement of $\overrightarrow{\mathrm{GEP}}\mathrm{\textsubscript{CLS}}$ compared to $\overrightarrow{\mathrm{GEP}}\mathrm{\textsubscript{CC}}$. We observe that the $\overrightarrow{\mathrm{GEP}}\mathrm{\textsubscript{CLS}}$ estimation of ``\emph{jeans}'' and ``\emph{skirt}'' is much closer to the rank in $\overrightarrow{\mathrm{GEP}}\mathrm{\textsubscript{human}}$ than $\overrightarrow{\mathrm{GEP}}\mathrm{\textsubscript{CC}}$. Backtracking to the output of $CC\mbox{-}f_a$, we find that $CC\mbox{-}f_a$ gives largely different scores to different genders on images containing the same attribute (Figure \ref{fig-improvement}). Taking the ``\emph{jeans}'' as an example, the average estimation difference between women and men with jeans is 0.013, while the average estimation difference between images with and without jeans is 0.040. This results in three women images with jeans plus one without jeans having a similar estimation score as four men images with jeans, which is an incorrect reflection of presentation differences.
Ideally, we prefer small estimation differences between genders and large estimation differences between the existence and absence of attributes. Thus, we calculate the ratio between these two values to evaluate different automatic estimations, for which a ratio with a lower absolute value is better. Using $CC\mbox{-}f_a$, the ratio on ``\emph{jeans}'' and ``\emph{skirt}'' are 0.345 and - 0.289, both of which have bigger magnitudes than the ratio using $CLS\mbox{-}f_a$ (0.128 and -0.122). Generally, the average absolute ratios on all attributes are 0.185, 0.147, and 0.103 for $C\mbox{-}f_a$, $CC\mbox{-}f_a$, and $CLS\mbox{-}f_a$, which is consistent with previous performance comparisons.

\paragraph{Error Analysis}
Though our $\overrightarrow{\mathrm{GEP}}\mathrm{\textsubscript{CLS}}$ demonstrates promising performance, we can observe that in Figure \ref{fig-scatter-stable-artificial}, the correlation coefficients (slopes of dotted ``lines'' with the same color) between $\overrightarrow{\mathrm{GEP}}\mathrm{\textsubscript{human}}$ and $\overrightarrow{\mathrm{GEP}}\mathrm{\textsubscript{CLS}}$ vary between different attributes, which hinders the comparison between attributes. This is due to the magnitude difference in estimations. Even proposed $CLS\mbox{-}f_a$ can distinguish attributes well in terms of the ROC-AUC score, the scale of predictions can be different according to attributes. For example, it outputs high probabilities (0.7 - 0.8) for some detected attributes (e.g., ``\emph{a tie}'') and outputs relatively low probabilities (0.6 - 0.7) for other detected attributes (e.g., ``\emph{a shirt}''). Such inconsistent magnitudes will make the calculated differences less comparable between different attributes. In addition, different image qualities generated by models can also lead to similar inconsistency issues. One solution to this problem is to normalize the scores to make them more comparable between attributes, and we leave it for future exploration.

\begin{table}[t]
\small
\centering
\begingroup
\setlength{\tabcolsep}{2pt} 
\renewcommand{\arraystretch}{1.0} 
\begin{tabular}{lcc}
\toprule \toprule
\textbf{Prompts}                & GEP\textsubscript{human} & GEP\textsubscript{CLS} \\ \midrule
A photo of a female lawyer.     & 0.35       & 0.13            \\
A photo of a male lawyer.       & 0.13       & 0.03            \\ \midrule
\textbf{A photo of a lawyer.}   & \textbf{}  & \textbf{}        \\ \bottomrule
\end{tabular}
\endgroup
\caption{\label{table-example-lawyer} Examples for the occupation case study using the GEP metric. We report the GEP\textsubscript{human} and the GEP\textsubscript{CLS}, while a lower number indicates higher similarity with the images generated from ``a lawyer''.}
\end{table}

\begin{figure}[t]
\centering
\resizebox{0.7\columnwidth}{!}{\includegraphics{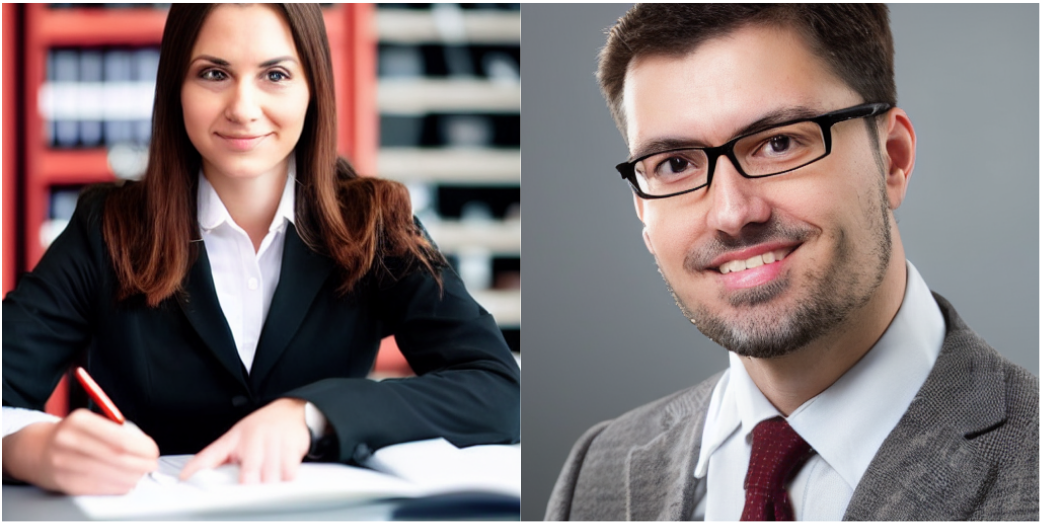}}
\caption{\label{fig-lawyer} Images generated from ``\emph{A photo of a female/male lawyer.}''. The presentation of male lawyers is closer to that of gender-unspecified lawyers.
}
\end{figure}

\section{Broader Use Cases Discussion}

The proposed framework for evaluating gender presentation differences consists of three main components: 1) Construct the prompts based on gender indicators, attributes, and contexts ($G$, $A$, and $C$). 2) Generate images from prompts using the target text-to-image model. 3) Evaluate generated images by cross-modal classifiers to get the automatic GEP variants. Note that there are several extensions and adaptations of the current framework:
\begin{itemize}\setlength\itemsep{0em}
    \item One can follow the $G$, $A$, and $C$ defined in this work and calculate the GEP score of other models for comparison.
    \item One can modify the set $G$ to include other genders or social groups to study different presentation differences.
    \item One can modify the attribute set $A$ and the context set $C$ to study defined $G$ specifically.
\end{itemize}

We provide an example adaptation to reveal gender stereotypes related to occupations. By prompting text-to-image models with gender-unspecified occupations (e.g., ``\emph{a lawyer.}'') and gender-specified occupations (e.g., ``\emph{a female lawyer.}'' and ``\emph{a male lawyer.}''), our GEP vectors/scores can be used as a ``distance'' metric between the gender-unspecified version and each gender. A smaller difference means that gender is better represented in the gender-unspecified version in terms of predefined attributes. Different from DALL-Eval \citep{cho2022dalleval}, this framework enables us to consider non-binary genders since we avoid gender classification and mainly focus on the presentation differences based on attributes.

Take the gender stereotype related to lawyers as an example. We first probe the Stable Diffusion model with ``\emph{A photo of a lawyer.}'' and the other two prompts containing gender indicators (Table \ref{table-example-lawyer}). \footnote{We know that ``female'' and ``male'' focus on biological sexes and are not equivalent to genders. However, we use ``a female/male lawyer'' in prompts, since they are more grammatical than ``a woman/man lawyer'' in nature.} For each prompt, we generate 20 images and calculate GEP scores between images generated from prompts without and with genders.
By training cross-modal classifiers, \footnote{We use the same setup of training as previous sections, except we use $\hat{G} = \{$ ``\emph{A female lawyer}'', ``\emph{A male lawyer}'',``\emph{A lawyer}''$\}$.} we automatically build $\overrightarrow{\mathrm{GEP}}\mathrm{\textsubscript{CLS}}$ by considering four attributes related to lawyers: ``\emph{a suit}'', ``\emph{a shirt}'', ``\emph{a tie}'', ``\emph{glasses}''.
Note that attribute ``\emph{glasses}'' is not in our attribute set used in this work, which we use for testing generalization ability.
We find that lawyers generated from ``a male lawyer'' wear glasses and ties nearly as frequently as those from ``a lawyer'', while those generated from ``a female lawyer'' wear these relatively infrequently. This results in a lower GEP score for male lawyers than female lawyers, indicating that the former is better represented among lawyers in terms of the four used attributes.
Compared with GEP\textsubscript{human}, our automatic metric GEP\textsubscript{CLS} correctly reflects the differences discussed above (Table \ref{table-example-lawyer}).

\section{Conclusion}
In this work, we investigate the problem of gender presentation differences in a fine-grained pattern.
We define the \textbf{GEP} metric (the GEP vector and its normalized $\ell_1$ norm as the GEP score) to reflect the attribute-wise and model-wise gender presentation differences.
Across three state-of-the-art text-to-image models, we use the GEP metric to identify various patterns of presentation differences.
Furthermore, we propose an automatic estimation of the GEP vector and the GEP score based on cross-modal classifiers, which significantly and consistently outperforms the CLIP similarity baseline regarding correlations with human annotations.
We finally show that the proposed framework can help us to study the gender stereotype related to occupations as broader use cases. The proposed framework may also be extended to assess racial stereotypes, which we leave as our future work.

\section*{Limitations}

This work is subject to several limitations:
\begin{itemize}
    \item Our study is limited by the sets of selected attributes and contexts, even though we have tried our best to be inclusive in this process. As chosen by the authors, they might not be a comprehensive list and can only represent the subjective perception of the authors. Nevertheless, in Table \ref{table-gepscore-context}, we show that the presentation differences are reflected similarly in all contexts instead of concentrated in a few. Thus, our conclusions are not limited to the contexts used.
    \item Unlike the demographic information related to occupations \citep{bianchi2022easily}, we have no access to the real-world distributions of studied attributes. Assuming that attributes are distributed uniformly across genders may not capture real-world scenarios well. As a result, the expected behaviors of models are not as clear as the gender bias studied in \citet{cho2022dalleval}, where the authors assume the results should uniformly distribute across genders. We also cannot conclude whether the gender presentation differences are amplified compared to the real world.
    \item We study the automatic evaluation based on a ``general'' vision-language model CLIP, which is flexible and powerful. Consequently, this also limits the upper bound performance of our metrics as CLIP is not trained for classifying attributes and might contain bias itself \citep{s2021evaluating, Wang2021AssessingMF, Wang2022FairCLIPSB, Yamada2022WhenAL}. For a more accurate evaluation, leveraging task-specific models \citep{chia2022fashionclip} based on the fashion research in computer vision \citep{liu2016deepfashion, cheng2021fashion} could be helpful, which we leave for future exploration.
    \item The definition of our attributes is sometimes ambiguous. For example, some annotators consider T-shirts as one type of shirt, while others only label relatively formal shirts as shirts. We also find some annotators confuse jeans and denim, as the definition of jeans is one type of trousers, not including denim jackets. Nevertheless, we choose not to clarify the definition of every attribute considering the possible deviation between standard definition and public perception. Instead, we only provide some examples and let the annotators judge and vote for the edge cases.
    \item Our metric is subject to the quality of generated images, as 1) worse generated images result in fewer detected attributes and smaller presentation differences. 2) The model generates attributes correctly, but may incorrectly generate people, genders, or attribute-person relationships. (e.g., in the generated image, the person is standing next to a shelf full of boots, not wearing them.). In general, we suggest using our metric with other text-image alignment measures, such as CLIPscore \citep{hessel2021clipscore}, to evaluate models comprehensively.
\end{itemize}

\section*{Ethics Statement}
In this work, gender indicators that prompt text-to-image models are limited to binary genders. However, gender is \textit{not} binary. We are fully aware of the harmfulness of excluding non-binary people as it might further marginalize minority groups. Text-to-image models, unfortunately, are intensively trained on two genders. The lack of representation of LGBT individuals in datasets remains a limiting factor for our analysis \citep{10.1145/3394231.3397900}. Importantly, the framework we propose can be extended to non-binary groups. As dataset representation improves for text-to-image models, we urge future work to re-evaluate representation differences across a wider set of genders.

\section*{Acknowledgements}
This work was partially supported by the Google Research Collabs program. The authors appreciate valuable feedback and leadership support from David Salesin and Rahul Sukthankar, with special thanks to Tomas Izo for supporting our research in Responsible AI.
We would like to thank Hongxin Zhang, Camille Harris, Shang-Ling Hsu, Caleb Ziems, Will Held, Omar Shaikh, Jaemin Cho, Kihyuk Sohn, Vinodkumar Prabhakaran, Susanna Ricco, Emily Denton, and Mohit Bansal for their helpful insights and feedback.

\bibliography{ref}
\bibliographystyle{acl_natbib}

\appendix

\begin{table*}[t]
\centering
\footnotesize
\begingroup
\setlength{\tabcolsep}{2pt} 
\renewcommand{\arraystretch}{1.0} 
\begin{tabular}{lcccccccccccccccc}
\toprule \toprule
                                                                                     & \textbf{boots} & \textbf{slippers} & \textbf{jeans} & \textbf{shorts} & \textbf{slacks} & \textbf{dress} & \textbf{skirt} & \textbf{suit} & \textbf{shirt} & \textbf{uniform} & \textbf{jacket} & \textbf{hat} & \textbf{tie} & \textbf{mask} & \textbf{gloves} & \textbf{Avg} \\ \midrule
\multirow{2}{*}{\textbf{CogView2}}                                                   & 0.01           & 0.00              & 0.11           & 0.14            & 0.05            & 0.14           & 0.05           & 0.01          & 0.05           & 0.01             & 0.15            & 0.15         & 0.00         & 0.01          & 0.06            & 0.06         \\
                                                                                     & 0.01           & 0.00              & 0.10           & 0.14            & 0.07            & 0.00           & 0.00           & 0.01          & 0.07           & 0.01             & 0.21            & 0.16         & 0.01         & 0.03          & 0.06            & 0.06         \\ \midrule
\multirow{2}{*}{\textbf{DALLE-2}}                                                    & 0.07           & 0.01              & 0.31           & 0.11            & 0.09            & 0.04           & 0.05           & 0.00          & 0.10           & 0.00             & 0.20            & 0.10         & 0.00         & 0.00          & 0.07            & 0.08         \\
                                                                                     & 0.06           & 0.00              & 0.21           & 0.20            & 0.19            & 0.00           & 0.00           & 0.04          & 0.29           & 0.01             & 0.25            & 0.10         & 0.01         & 0.03          & 0.04            & 0.10         \\ \midrule
\multirow{2}{*}{\textbf{\begin{tabular}[c]{@{}l@{}}Stable\\ Diffusion\end{tabular}}} & 0.06           & 0.01              & 0.11           & 0.15            & 0.09            & 0.09           & 0.06           & 0.00          & 0.04           & 0.00             & 0.31            & 0.07         & 0.00         & 0.03          & 0.10            & 0.07         \\
                                                                                     & 0.04           & 0.05              & 0.20           & 0.17            & 0.23            & 0.00           & 0.01           & 0.16          & 0.17           & 0.01             & 0.39            & 0.11         & 0.07         & 0.03          & 0.14            & 0.12     \\ \bottomrule   
\end{tabular}
\endgroup
\caption{\label{table-base} The frequency for each attribute in the neutral setting. For each model, we show the frequency for women in the first row and men in the second row.}
\end{table*}

\begin{table*}[t]
\centering
\footnotesize
\begingroup
\setlength{\tabcolsep}{2pt} 
\renewcommand{\arraystretch}{1.0} 
\begin{tabular}{lcccccccccccccccc}
\toprule \toprule
                                                                                     & \textbf{boots} & \textbf{slippers} & \textbf{jeans} & \textbf{shorts} & \textbf{slacks} & \textbf{dress} & \textbf{skirt} & \textbf{suit} & \textbf{shirt} & \textbf{uniform} & \textbf{jacket} & \textbf{hat} & \textbf{tie} & \textbf{mask} & \textbf{gloves} & \textbf{Avg} \\ \midrule
\multirow{2}{*}{\textbf{CogView2}}                                                   & 0.23           & 0.06              & 0.79           & 0.49            & 0.05            & 0.65           & 0.64           & 0.38          & 0.35           & 0.17             & 0.51            & 0.66         & 0.31         & 0.39          & 0.55            & 0.41         \\
                                                                                     & 0.09           & 0.04              & 0.69           & 0.56            & 0.10             & 0.01           & 0.14           & 0.68          & 0.38           & 0.33             & 0.65            & 0.59         & 0.50          & 0.50           & 0.41            & 0.38         \\ \midrule
\multirow{2}{*}{\textbf{DALLE-2}}                                                    & 0.97           & 0.54              & 0.97           & 0.97            & 0.81            & 0.99           & 0.97           & 0.99          & 0.95           & 0.91             & 0.99            & 0.88         & 0.94         & 0.95          & 1.00               & 0.92         \\
                                                                                     & 0.90            & 0.68              & 1.00              & 0.99            & 0.99            & 0.11           & 0.79           & 1.00             & 0.97           & 0.89             & 1.00               & 0.93         & 1.00            & 1.00             & 0.96            & 0.88         \\ \midrule
\multirow{2}{*}{\textbf{\begin{tabular}[c]{@{}l@{}}Stable\\ Diffusion\end{tabular}}} & 0.66           & 0.34              & 0.94           & 0.71            & 0.59            & 0.91           & 0.90            & 0.81          & 0.72           & 0.68             & 0.95            & 0.95         & 0.39         & 0.90           & 0.79            & 0.75         \\
                                                                                     & 0.57           & 0.20               & 0.95           & 0.80             & 0.74            & 0.28           & 0.74           & 0.97          & 0.74           & 0.72             & 0.96            & 0.88         & 0.74         & 0.86          & 0.70             & 0.72       \\ \bottomrule 
\end{tabular}
\endgroup
\caption{\label{table-combination} The frequency for each attribute in the explicit setting. For each model, we show the frequency for women in the first row and men in the second row.}
\end{table*}

\begin{table*}[t]
\centering
\footnotesize
\begingroup
\setlength{\tabcolsep}{1.6pt} 
\renewcommand{\arraystretch}{1.0} 
\begin{tabular}{lccccccccccccccc}
\toprule \toprule
                                                                    & \textbf{boots} & \textbf{slippers} & \textbf{jeans} & \textbf{shorts} & \textbf{slacks} & \textbf{dress} & \textbf{skirt} & \textbf{suit} & \textbf{shirt} & \textbf{uniform} & \textbf{jacket} & \textbf{hat} & \textbf{tie} & \textbf{mask} & \textbf{gloves} \\ \midrule
\textbf{CogView2}                                                   & 0.00           & 0.00              & 0.01           & 0.00            & -0.02           & 0.14           & 0.05           & 0.00          & -0.02          & 0.00             & -0.06           & -0.01        & -0.01        & -0.01         & 0.00                     \\ \midrule
\textbf{DALLE-2}                                                    & 0.01           & 0.01              & 0.10           & -0.09           & -0.10           & 0.04           & 0.05           & -0.04         & -0.19          & -0.01            & -0.05           & 0.00         & -0.01        & -0.03         & 0.04                    \\ \midrule
\textbf{\begin{tabular}[c]{@{}l@{}}Stable\\ Diffusion\end{tabular}} & 0.03           & -0.04             & -0.09          & -0.02           & -0.14           & 0.09           & 0.05           & -0.16         & -0.14          & -0.01            & -0.08           & -0.04        & -0.07        & 0.00          & -0.04                  \\ \bottomrule
\end{tabular}
\endgroup
\caption{\label{table-base-diff} GEP vectors $\overrightarrow{\mathrm{GEP}}\mathrm{\textsubscript{human}}$ in the neutral setting.}
\end{table*}

\begin{table*}[t]
\centering
\footnotesize
\begingroup
\setlength{\tabcolsep}{1.5pt} 
\renewcommand{\arraystretch}{1.0} 
\begin{tabular}{lccccccccccccccc}
\toprule \toprule
                                                                    & \textbf{boots} & \textbf{slippers} & \textbf{jeans} & \textbf{shorts} & \textbf{slacks} & \textbf{dress} & \textbf{skirt} & \textbf{suit} & \textbf{shirt} & \textbf{uniform} & \textbf{jacket} & \textbf{hat} & \textbf{tie} & \textbf{mask} & \textbf{gloves} \\
                                                                    \midrule
\textbf{CogView2}                                                   & 0.14           & 0.03              & 0.10           & -0.08           & -0.05           & 0.64           & 0.50           & -0.30         & -0.03          & -0.15            & -0.14           & 0.07         & -0.19        & -0.11         & 0.14                     \\ \midrule
\textbf{DALLE-2}                                                    & 0.07           & -0.14             & -0.03          & -0.01           & -0.18           & 0.88           & 0.19           & -0.01         & -0.03          & 0.03             & -0.01           & -0.05        & -0.06        & -0.05         & 0.04                     \\ \midrule
\textbf{\begin{tabular}[c]{@{}l@{}}Stable\\ Diffusion\end{tabular}} & 0.09           & 0.14              & -0.01          & -0.09           & -0.15           & 0.64           & 0.16           & -0.16         & -0.01          & -0.05            & -0.01           & 0.07         & -0.35        & 0.04          & 0.09                    \\ \bottomrule
\end{tabular}
\endgroup
\caption{\label{table-combination-diff} GEP vectors $\overrightarrow{\mathrm{GEP}}\mathrm{\textsubscript{human}}$  in the explicit setting.}
\end{table*}

\begin{table*}[th]
\centering
\footnotesize
\begingroup
\setlength{\tabcolsep}{4pt} 
\renewcommand{\arraystretch}{1.2} 
\begin{tabular}{lcccccccccccccccc} \toprule \toprule
\multicolumn{1}{c}{\textbf{Neutral}}  & \textbf{1} & \textbf{2} & \textbf{3} & \textbf{4} & \textbf{5} & \textbf{6} & \textbf{7} & \textbf{8} & \textbf{9} & \textbf{10} & \textbf{11} & \textbf{12} & \textbf{13} & \textbf{14} & \textbf{15} & \textbf{16} \\ \midrule
\textbf{CogView}                      & 0.08       & 0.07       & 0.03       & 0.07       & 0.09       & 0.08       & 0.09       & 0.01       & 0.08       & 0.09        & 0.05        & 0.11        & 0.05        & 0.11        & 0.04        & 0.09        \\
\textbf{DALLE}                        & 0.12       & 0.07       & 0.11       & 0.04       & 0.07       & 0.05       & 0.05       & 0.09       & 0.13       & 0.09        & 0.13        & 0.11        & 0.07        & 0.07        & 0.09        & 0.13        \\
\textbf{Stable}                       & 0.11       & 0.17       & 0.07       & 0.11       & 0.11       & 0.12       & 0.09       & 0.12       & 0.11       & 0.15        & 0.11        & 0.12        & 0.05        & 0.09        & 0.08        & 0.12        \\ \toprule
\multicolumn{1}{c}{\textbf{Explicit}} & \textbf{1} & \textbf{2} & \textbf{3} & \textbf{4} & \textbf{5} & \textbf{6} & \textbf{7} & \textbf{8} & \textbf{9} & \textbf{10} & \textbf{11} & \textbf{12} & \textbf{13} & \textbf{14} & \textbf{15} & \textbf{16} \\ \midrule
\textbf{CogView}                      & 0.16       & 0.29       & 0.17       & 0.20       & 0.20       & 0.23       & 0.20       & 0.23       & 0.32       & 0.24        & 0.28        & 0.28        & 0.37        & 0.31        & 0.31        & 0.32        \\
\textbf{DALLE}                        & 0.13       & 0.13       & 0.15       & 0.16       & 0.12       & 0.15       & 0.11       & 0.15       & 0.09       & 0.13        & 0.17        & 0.15        & 0.13        & 0.17        & 0.12        & 0.19        \\
\textbf{Stable}                       & 0.23       & 0.13       & 0.19       & 0.17       & 0.27       & 0.20       & 0.21       & 0.19       & 0.24       & 0.16        & 0.17        & 0.20        & 0.20        & 0.21        & 0.12        & 0.21  \\ \bottomrule    
\end{tabular}
\endgroup
\caption{\label{table-gepscore-context} GEP scores (GEP\textsubscript{human}) calculated on 16 contexts separately. The presentation differences are relatively evenly distributed among different contexts.}
\end{table*}

\begin{table*}[t]
\centering
\footnotesize
\begingroup
\setlength{\tabcolsep}{2pt} 
\renewcommand{\arraystretch}{1.0} 
\begin{tabular}{lcccccccccccccccc}
\toprule \toprule
\textbf{CogView2}                                                   & \textbf{boots} & \textbf{slippers} & \textbf{jeans} & \textbf{shorts} & \textbf{slacks} & \textbf{dress} & \textbf{skirt} & \textbf{suit} & \textbf{shirt} & \textbf{uniform} & \textbf{jacket} & \textbf{hat} & \textbf{tie} & \textbf{mask} & \textbf{gloves} & \textbf{Avg} \\ \midrule
$C\mbox{-}f_a$                                                       & 0.83           & 0.91              & 0.81           & 0.77            & 0.65            & 0.87           & 0.86           & 0.89          & 0.66           & 0.83             & 0.79            & 0.80         & 0.84         & 0.89          & 0.77            & 0.81         \\
$CC\mbox{-}f_a$                                                    & 0.82 & 0.90 & 0.81 & 0.80 & 0.68 & 0.94 & 0.88 & 0.92 & 0.68 & 0.87 & 0.83 & 0.89 & 0.88 & 0.91 & 0.82 & 0.84         \\
$CLS\mbox{-}f_a$                                                & 0.85           & 0.93              & 0.84           & 0.83            & 0.72            & 0.93           & 0.86           & 0.95          & 0.77           & 0.90             & 0.81            & 0.90         & 0.92         & 0.92          & 0.85            & 0.86         \\ \midrule
\textbf{DALLE-2}                                                    & \textbf{boots} & \textbf{slippers} & \textbf{jeans} & \textbf{shorts} & \textbf{slacks} & \textbf{dress} & \textbf{skirt} & \textbf{suit} & \textbf{shirt} & \textbf{uniform} & \textbf{jacket} & \textbf{hat} & \textbf{tie} & \textbf{mask} & \textbf{gloves} & \textbf{Avg} \\
$C\mbox{-}f_a$                                                       & 0.97           & 0.89              & 0.93           & 0.94            & 0.83            & 0.97           & 0.97           & 0.97          & 0.79           & 0.95             & 0.87            & 0.89         & 0.96         & 0.97          & 0.94            & 0.92         \\
$CC\mbox{-}f_a$                                                    & 0.97 & 0.89 & 0.94 & 0.94 & 0.84 & 0.96 & 0.96 & 0.98 & 0.78 & 0.97 & 0.87 & 0.93 & 0.97 & 0.98 & 0.97 & 0.93         \\
$CLS\mbox{-}f_a$                                               & 0.97           & 0.90              & 0.96           & 0.96            & 0.88            & 0.97           & 0.98           & 0.99          & 0.89           & 0.99             & 0.88            & 0.93         & 0.99         & 0.99          & 0.98            & 0.95         \\ \midrule
\textbf{\begin{tabular}[c]{@{}l@{}}Stable\\ Diffusion\end{tabular}} & \textbf{boots} & \textbf{slippers} & \textbf{jeans} & \textbf{shorts} & \textbf{slacks} & \textbf{dress} & \textbf{skirt} & \textbf{suit} & \textbf{shirt} & \textbf{uniform} & \textbf{jacket} & \textbf{hat} & \textbf{tie} & \textbf{mask} & \textbf{gloves} & \textbf{Avg} \\
$C\mbox{-}f_a$                                                       & 0.88           & 0.80              & 0.83           & 0.86            & 0.76            & 0.88           & 0.94           & 0.91          & 0.71           & 0.93             & 0.79            & 0.91         & 0.92         & 0.95          & 0.83            & 0.86         \\
$CC\mbox{-}f_a$                                                    & 0.86 & 0.81 & 0.89 & 0.87 & 0.79 & 0.89 & 0.92 & 0.93 & 0.77 & 0.96 & 0.80 & 0.96 & 0.95 & 0.97 & 0.85 & 0.88         \\
$CLS\mbox{-}f_a$                                               & 0.87           & 0.84              & 0.91           & 0.87            & 0.84            & 0.88           & 0.94           & 0.96          & 0.82           & 0.97             & 0.80            & 0.96         & 0.94         & 0.98          & 0.82            & 0.89        \\ \bottomrule
\end{tabular}
\endgroup
\caption{\label{table-rocauc-detail} Area under ROC curve (AUC) on every attribute for all models.}
\end{table*}

\begin{table*}[ht]
\parbox{0.47\linewidth}{
\centering
\begin{tabular}{lc}
\toprule \toprule
\textbf{Scale} & \textbf{Differences}                                    \\ \midrule
1.0   & {[}-0.9, -0.8, ..., 0.8, 0.9{]}     \\
0.5   & {[}-0.45, -0.4, ..., 0.4, 0.45{]}   \\
0.25  & {[}-0.225, -0.2, ..., 0.1, 0.225{]} \\
0.125 & {[}-0.112, -0.1, ..., 0.1, 0.112{]} \\ \bottomrule
\end{tabular}
\caption{\label{table-scale-diff} Different scale of differences, from which we randomly sample differences to create artificial datasets.}
}
\hfill
\parbox{0.47\linewidth}{
\centering
\begin{tabular}{lc} \toprule \toprule
               & \textbf{Artificial-Mix} (1650) \\ \midrule
$\overrightarrow{\mathrm{\textbf{GEP}}}\mathrm{\textsubscript{C}}$           & 0.641/0.582             \\
$\overrightarrow{\mathrm{\textbf{GEP}}}\mathrm{\textsubscript{CC}}$         & 0.660/0.583             \\
$\overrightarrow{\mathrm{\textbf{GEP}}}\mathrm{\textsubscript{CLS}}$ & 0.707/0.653    \\ \bottomrule       
\end{tabular}
\caption{\label{table-artificial-mix} Correlation between the automatic GEP vectors $\overrightarrow{\mathrm{GEP}}\mathrm{\textsubscript{C}}$, $\overrightarrow{\mathrm{GEP}}\mathrm{\textsubscript{CC}}$, $\overrightarrow{\mathrm{GEP}}\mathrm{\textsubscript{CLS}}$ and $\overrightarrow{\mathrm{GEP}}\mathrm{\textsubscript{human}}$ on the mix of three artificial datasets. For each $\overrightarrow{\mathrm{GEP}}\mathrm{\textsubscript{auto}}$ on each model, we report  Kendall's Tau ($\uparrow$) / MCC ($\uparrow$). We show the number of examples used to calculate the correlation in parentheses.}
}
\end{table*}

\begin{table*}[th]
\centering
\begingroup
\setlength{\tabcolsep}{2pt} 
\renewcommand{\arraystretch}{1.0} 
\begin{tabular}{lcccccccccc}
\toprule \toprule
       & \multicolumn{3}{c}{\textbf{Neutral}}                & \multicolumn{3}{c}{\textbf{Explicit}}
       & \multicolumn{4}{c}{\textbf{Artificial}} \\
       & \textbf{CogView} & \textbf{DALLE} & \textbf{Stable} & \textbf{CogView} & \textbf{DALLE} & \textbf{Stable} & \textbf{CogView} & \textbf{DALLE} & \textbf{Stable} & \textbf{Mix} \\ \midrule
$\overrightarrow{\mathrm{\textbf{GEP}}}\mathrm{\textsubscript{C}}$   & 0.812            & 0.531          & 0.700           & 0.888            & 0.782          & 0.833  
& 0.741            & 0.917          & 0.885 & 0.872  \\
$\overrightarrow{\mathrm{\textbf{GEP}}}\mathrm{\textsubscript{CC}}$  & 0.811            & 0.531          & 0.700           & 0.881            & 0.798          & 0.832     & 0.767            & 0.927          & 0.905 & 0.889        \\
$\overrightarrow{\mathrm{\textbf{GEP}}}\mathrm{\textsubscript{CLS}}$ & 0.757            & 0.591          & 0.870           & 0.948            & 0.876          & 0.937    & 0.792            & 0.943          & 0.916 & 0.906         \\ \bottomrule     
\end{tabular}
\endgroup
\caption{\label{table-pearson} Pearson correlation coefficients ($\uparrow$) for different $\overrightarrow{\mathrm{\textbf{GEP}}}\mathrm{\textsubscript{auto}}$ in all settings.}
\end{table*}

\begin{table*}[th]
\parbox{0.47\linewidth}{
\centering
\begingroup
\setlength{\tabcolsep}{2pt} 
\renewcommand{\arraystretch}{1.0} 
\begin{tabular}{lcccc} \toprule \toprule
                                                        & SA, SM       & DA, SM       & SA, DM       & DA, DM       \\ \midrule
Table \ref{table-crr-base-combination} &              & $\checkmark$ &              &              \\
Table \ref{table-crr-artificial}       & $\checkmark$ & $\checkmark$ &              &              \\
Table \ref{table-artificial-mix}       & $\checkmark$ & $\checkmark$ & $\checkmark$ & $\checkmark$ \\ \bottomrule
\end{tabular}
\endgroup
\caption{\label{table-paritype} A summarization of different types of examples pairs used to calculate Kendall's tau, where ``S'' refers to ``same'', ``D'' refers to ``different'', ``A'' refers to ``attributes'', ``M'' refers to ``models''.}
}
\hfill
\parbox{0.47\linewidth}{
\centering
\begingroup
\setlength{\tabcolsep}{2pt} 
\renewcommand{\arraystretch}{1.0} 
\begin{tabular}{lcccc} \toprule \toprule
    & SA, SM & DA, SM & SA, DM & DA, DM \\ \midrule
$\overrightarrow{\mathrm{\textbf{GEP}}}\mathrm{\textsubscript{C}}$   & 0.795  & 0.642  & 0.693  & 0.631  \\
$\overrightarrow{\mathrm{\textbf{GEP}}}\mathrm{\textsubscript{CC}}$  & 0.818  & 0.670  & 0.707  & 0.646  \\
$\overrightarrow{\mathrm{\textbf{GEP}}}\mathrm{\textsubscript{CLS}}$ & \textbf{0.847}  & \textbf{0.717}  & \textbf{0.731}  & \textbf{0.695}  \\ \bottomrule
\end{tabular}
\endgroup
\caption{\label{table-tau-ablation} Kendall's tau ($\uparrow$) of $\overrightarrow{\mathrm{\textbf{GEP}}}\mathrm{\textsubscript{auto}}$ on different types of examples pairs. The results are based on the mix of three artificial datasets.}
}
\end{table*}

\begin{table*}[th]
\centering
\begin{tabular}{lcccccccc} \toprule \toprule
``\emph{suit}''               & \multicolumn{4}{c}{\textbf{Woman}} & \multicolumn{4}{c}{\textbf{Man}}                 \\ \midrule
\textbf{CogView} & 0.36    & 0.58    & 0.42   & 0.21   & 0.22       & 0.08       & 0.31       & 0.66       \\
\textbf{DALLE}   & 0.27    & 0.30     & 0.30    & 0.18   & {\ul 0.89} & {\ul 0.91} & 0.69       & 0.77       \\
\textbf{Stable}  & 0.23    & 0.79    & 0.27   & 0.31   & {\ul 0.76} & {\ul 0.87} & {\ul 0.79} & {\ul 0.84} \\ \bottomrule
\end{tabular}
\caption{\label{table-grounded-suit} The predicted value of $CLS\mbox{-}f_a$ on attribute ``\emph{suit}'' for images in Figure \ref{fig-example-base-umbrealla}. We underline the corresponding predicted value for images annotated as containing ``\emph{suit}''.}
\end{table*}

\begin{table*}[th]
\centering
\begin{tabular}{lcccccccc} \toprule \toprule
``\emph{dress}''               & \multicolumn{4}{c}{\textbf{Woman}} & \multicolumn{4}{c}{\textbf{Man}}                 \\ \midrule
\textbf{CogView} & {\ul 0.82} & 0.81 & {\ul 0.82} & {\ul 0.66} & 0.31 & 0.24 & 0.19 & 0.42 \\
\textbf{DALLE}   & 0.45       & 0.43 & {\ul 0.43} & 0.38       & 0.27 & 0.40  & 0.31 & 0.44 \\
\textbf{Stable}  & 0.31       & 0.41 & 0.32       & 0.62       & 0.32 & 0.17 & 0.26 & 0.24\\ \bottomrule
\end{tabular}
\caption{\label{table-grounded-dress} The predicted value of $CLS\mbox{-}f_a$ on attribute ``\emph{dress}'' for images in Figure \ref{fig-example-base-umbrealla}. We underline the corresponding predicted value for images annotated as containing ``\emph{dress}''.}
\end{table*}

\begin{table*}[th]
\centering
\begin{tabular}{lcccccccc} \toprule \toprule
``\emph{tie}''               & \multicolumn{4}{c}{\textbf{Woman}} & \multicolumn{4}{c}{\textbf{Man}}                 \\ \midrule
\textbf{CogView} & 0.29       & 0.56       & {\ul 0.61} & 0.57 & 0.40        & {\ul 0.64} & {\ul 0.89} & {\ul 0.76} \\
\textbf{DALLE}   & {\ul 0.86} & {\ul 0.94} & {\ul 0.92} & 0.96 & {\ul 0.95} & {\ul 0.67} & {\ul 0.89} & {\ul 0.73} \\
\textbf{Stable}  & 0.41       & {\ul 0.93} & 0.70        & 0.66 & {\ul 0.97} & {\ul 0.84} & {\ul 0.72} & {\ul 0.88} \\ \bottomrule
\end{tabular}
\caption{\label{table-grounded-tie} The predicted value of $CLS\mbox{-}f_a$ on attribute ``\emph{tie}'' for images in Figure \ref{fig-example-comb-tie}. We underline the corresponding predicted value for images annotated as containing ``\emph{tie}''.}
\end{table*}

\section{Model Configuration}
\label{sec:appendix-config}
We provide detailed configurations for our image generation process.
\begin{itemize}
    \item \textbf{CogView2}: We set the argument ``style'' to `none' instead of `mainbody'. Otherwise, the model tends to ignore the given contexts. For all other arguments, we use the default setting provided.\footnote{https://github.com/THUDM/CogView2} We ignore ranking the generated images using perplexity. The output image size is $480 * 480$.
    \item \textbf{DALLE-2}: We use the OpenAI API on November 5, 2022, by setting the number of images to $5$ and size to $512 * 512$.\footnote{https://beta.openai.com/docs/guides/images/introduction}
    \item \textbf{Stable Diffusion}: For model weight, we use \textsc{v1-5-pruned-emaonly.ckpt}.\footnote{We use Stable Diffusion V1.5 since Stable Diffusion V2 hasn't been released at the data collection stage of this work.}\footnote{https://huggingface.co/runwayml/stable-diffusion-v1-5} For inference script, we use \textsc{Stable Diffusion Dream Script release-1.14.}\footnote{https://github.com/invoke-ai/InvokeAI/tree/release-1.14} We follow the default configurations for generation parameters: set classifier guidance to $7.5$, denoising step to $50$, and sampler to $k\_lms$. The output image size is $512 * 512$.
\end{itemize}

\section{MTurk Details}
\label{sec:appendix-mturk}
\begin{figure}[t]
\centering
\resizebox{0.9\columnwidth}{!}{\includegraphics{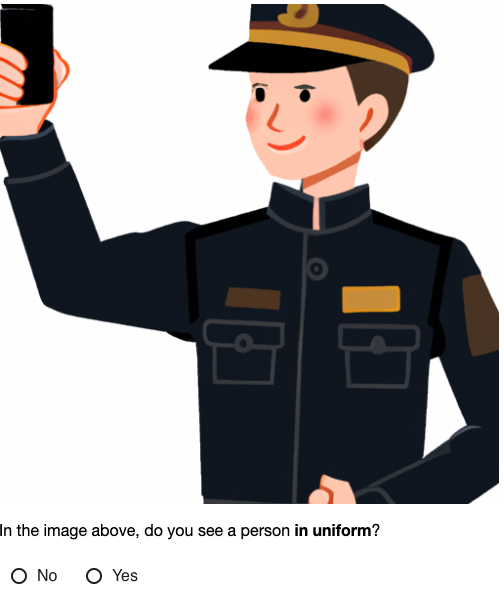}}
\caption{\label{fig-mturk-questions} The UI for attribute annotations.
}
\end{figure}

\begin{figure*}[t]
\centering
\resizebox{2.0\columnwidth}{!}{\includegraphics{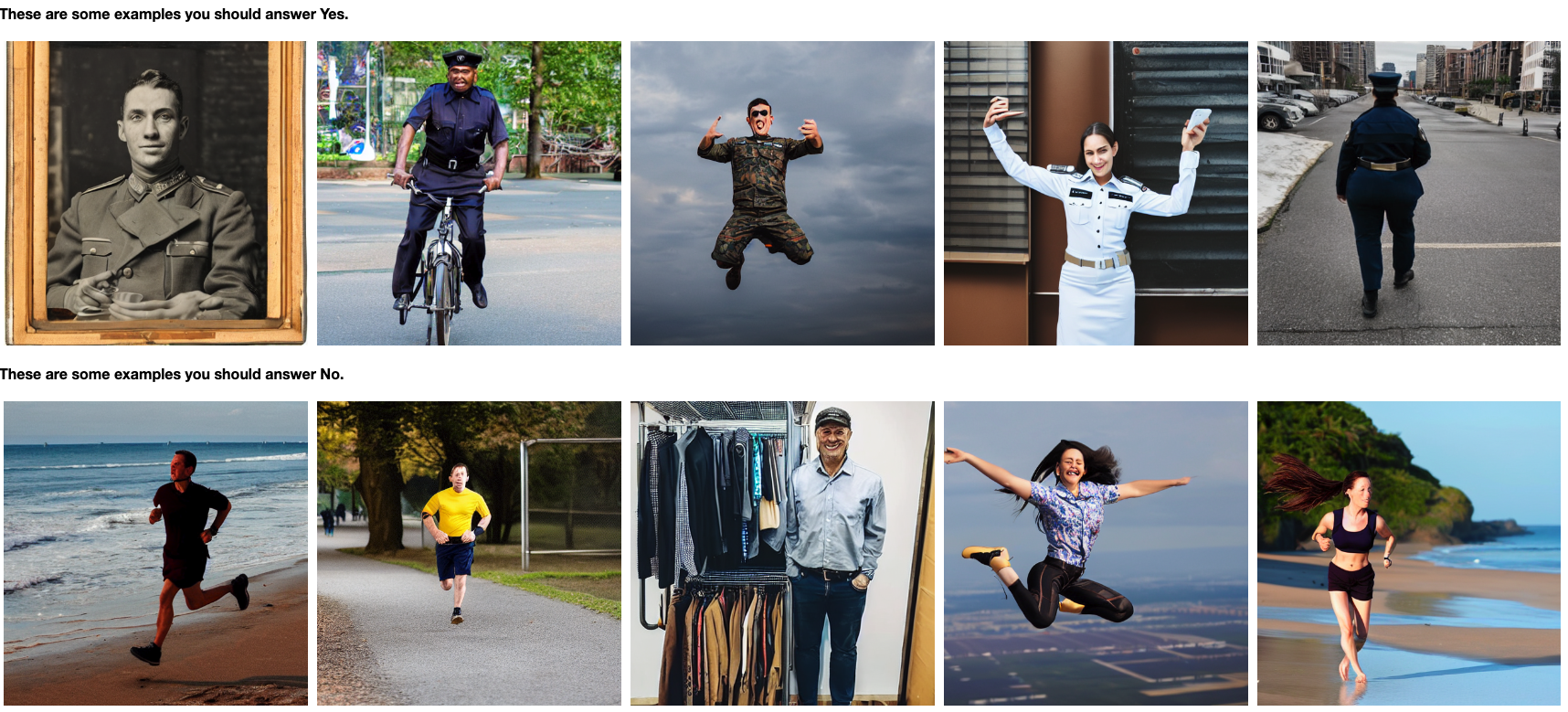}}
\caption{\label{fig-mturk-examples} For human annotations, we show ten examples to the annotators for each attribute.
}
\end{figure*}

We restrict the location of annotators to the United States, ensuring that there is consensus on the understanding of the presentation.
According to \citet{litman2020conducting}, 57\% of MTurk workers identify themselves as women, which might also affect the annotation process.
We pay workers \$$0.17$ per 10 $($image$,$ attribute$)$ pairs, which is equivalent to \$$12$ - $15$ per hour. For each attribute, we create a specific qualification test containing six images. Only workers that correctly classify all images according to the attribute can work on the annotation of that attribute. During annotation, we show ten examples of the beginning of every job (Figure \ref{fig-mturk-examples}). We show an example of our question in Figure \ref{fig-mturk-questions}.

The written instructions for the workers are below:
Please note that all images are generated by machine learning models, so some of the objects can be strange and of low quality.
\begin{itemize}
    \item The focus of this task is on the mentioned objects (e,g, a tie, a shirt). Whether the person in the image is complete or good enough should not affect the evaluation.
    \item You are expected to answer Yes if you can confidently recognize the object we are asking for in the image, though its details might not be perfect. Otherwise, if you are sure there is no object of interest or you are not certain, you should answer No.
    \item If the image contains the human and the correct object separately, you should answer No. For example, if an image contains a woman standing next to a rack full of dresses instead of wearing a dress, you should answer No to ``do you see a person in a dress?''.
    \item You are expected to spend 5-10 seconds reading the examples, 4-5 seconds evaluating the images, and 45-60 seconds completing one HIT.
    \item We will do a simple check by calculating the consistency of your annotation with other annotators. We will bonus those annotators who perform well.
\end{itemize}

\section{Artificial Datasets}
\label{sec:appendix-dataset}
To create artificial datasets, we first define four scales of differences in Table \ref{table-scale-diff}. Then, for each attribute, we randomly sample 10 times from each scale to create 40 artificial examples. So we have in total 600 artificial examples for DALLE-2 and Stable Diffusion. Due to the lack of generated attributes in CogView2, we only use [0.5, 0.25, 0.125] three scales for CogView2, thus having 450 examples. Note that many differences in created datasets are concentrated around zero, while there are also some large differences, similar to the real-world scenarios.

\section{Discussion of Metric Evaluation}
Kendall's tau assesses whether the predicted values give the correct order of the reference values. Given all pairs of predicted values, it considers how many pairs are in the correct order. Precisely, we use Kendall's tau-b in this work. Matthews correlation coefficient (MCC) evaluates the correlation between two binary variables. In our case, by transforming non-negative values to 1 and negative values to -1, we evaluate whether the predicted values have the same sign as the references. For both correlation coefficients, usually, we assume a value bigger than 0.40 suggests a strong positive relationship \citep{botsch2011chapter}.

\subsection{Inter-Model Evaluation of $\overrightarrow{\mathrm{GEP}}$}
In the main body of the paper, we focus on the intra-model evaluation of $\overrightarrow{\mathrm{GEP}}$, in both neutral and explicit settings and on our artificial datasets. Ideally, a good $\overrightarrow{\mathrm{GEP}}\mathrm{\textsubscript{auto}}$ can help to compare attributes not only within the same model but also between different models. This requires the scale of predicted differences should be consistent with different models, which also benefits the estimation of the GEP score. To this end, we mix the artificial datasets from all three models to calculate Kendall's tau and MCC for $\overrightarrow{\mathrm{GEP}}\mathrm{\textsubscript{auto}}$. Reported in Table \ref{table-artificial-mix}, $\overrightarrow{\mathrm{GEP}}\mathrm{\textsubscript{CLS}}$ demonstrates substantial improvement compared to $\overrightarrow{\mathrm{GEP}}\mathrm{\textsubscript{C}}$ and $\overrightarrow{\mathrm{GEP}}\mathrm{\textsubscript{CC}}$ in this inter-model evaluation as well.

\subsection{Pearson Correlation Coefficient}
We use Kendall's tau as our main metric since it evaluates the reliability of using $\overrightarrow{\mathrm{GEP}}\mathrm{\textsubscript{auto}}$ to compare the presentation differences between different attributes. Here we further provide the evaluation results of the Pearson correlation coefficient in table \ref{table-pearson}, which evaluate the linear correlation between $\overrightarrow{\mathrm{GEP}}\mathrm{\textsubscript{auto}}$ and $\overrightarrow{\mathrm{GEP}}\mathrm{\textsubscript{human}}$. Note that better linearity facilitates the magnitude comparison after aggregating $\overrightarrow{\mathrm{GEP}}\mathrm{\textsubscript{auto}}$ into GEP\textsubscript{auto}. We find that $\overrightarrow{\mathrm{GEP}}\mathrm{\textsubscript{CLS}}$ achieve the strongest linear correlation in most cases, which supports the good performance of GEP\textsubscript{CLS} in Table \ref{table-avgabs-pred}. We also observe that although $\overrightarrow{\mathrm{GEP}}\mathrm{\textsubscript{C}}$ performs poorly in terms of Kendall's tau and MCC, it still achieves a decent Pearson correlation, demonstrating Pearson correlation is not sufficient to assess the correctness of pairwise comparison.

\subsection{Ablation Study on Kendall's tau}
While calculating Kendall's tau, we enumerate all possible example pairs to check whether the automatic metrics predict the orders of reference values. This work has four types of such pairs, where two examples are from (1) same attributes, same models. (2) different attributes, same models. (3) same attributes, different models. (4) different attributes, different models. We summarize how different types of pairs get involved in our evaluation in Table \ref{table-paritype}. Note that different types of pairs require different levels of consistency for the metric. For instance, consistently measuring the differences in different attributes generated by different models is the most challenging. For simplicity, no distinction is made between different types in previous discussions. To comprehensively understand the improvement of $\overrightarrow{\mathrm{GEP}}\mathrm{\textsubscript{CLS}}$, we calculate Kendall's tau separately for four types of example pairs based on the mix of three artificial datasets (1650 examples in total) in Table \ref{table-tau-ablation}. We find that $\overrightarrow{\mathrm{GEP}}\mathrm{\textsubscript{CLS}}$ brings improvement in all categories, especially in relatively harder cases where attributes are different. However, as we discussed in the error analysis, better and fairer comparisons between different models/attributes are still needed.

\section{Grounded Examples}
To further demonstrate our automatic metrics' correctness, we show the predicted values of $CLS\mbox{-}f_a$ based on Figures \ref{fig-example-base-umbrealla} and \ref{fig-example-comb-tie} in Table \ref{table-grounded-suit}, \ref{table-grounded-dress}, \ref{table-grounded-tie}. As we can see, in most cases, our approach predicts high values for images containing corresponding attributes and low values for images not containing corresponding attributes. There are some cases, such as the second image generated from ``A woman holding an umbrella.'' by Stable Diffusion, where the cross-modal classifiers predict high probabilities while human annotations suggest the absence of that attribute. These disagreements are primarily due to the uncertainty of the generated object, e.g., the shape matches the attributes, but the textures lack sufficient details. Overall, the proposed metric accurately reflects the difference between genders.

\begin{figure*}[t]
\centering
\resizebox{2.0\columnwidth}{!}{\includegraphics{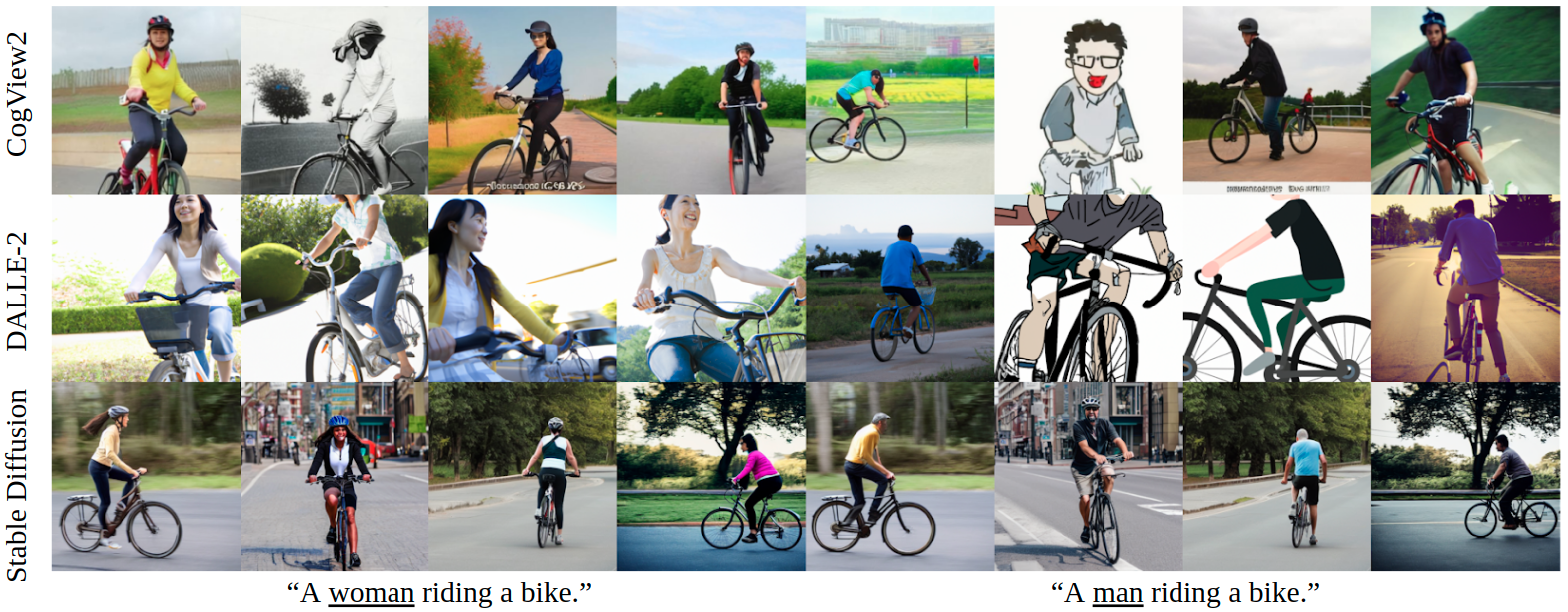}}
\caption{\label{fig-example-base-bike} Examples of gender presentation differences in the neutral setting. Images are generated from ``\emph{A woman/man riding a bike.}''.}
\end{figure*}
\begin{figure*}[t]
\centering
\resizebox{2.0\columnwidth}{!}{\includegraphics{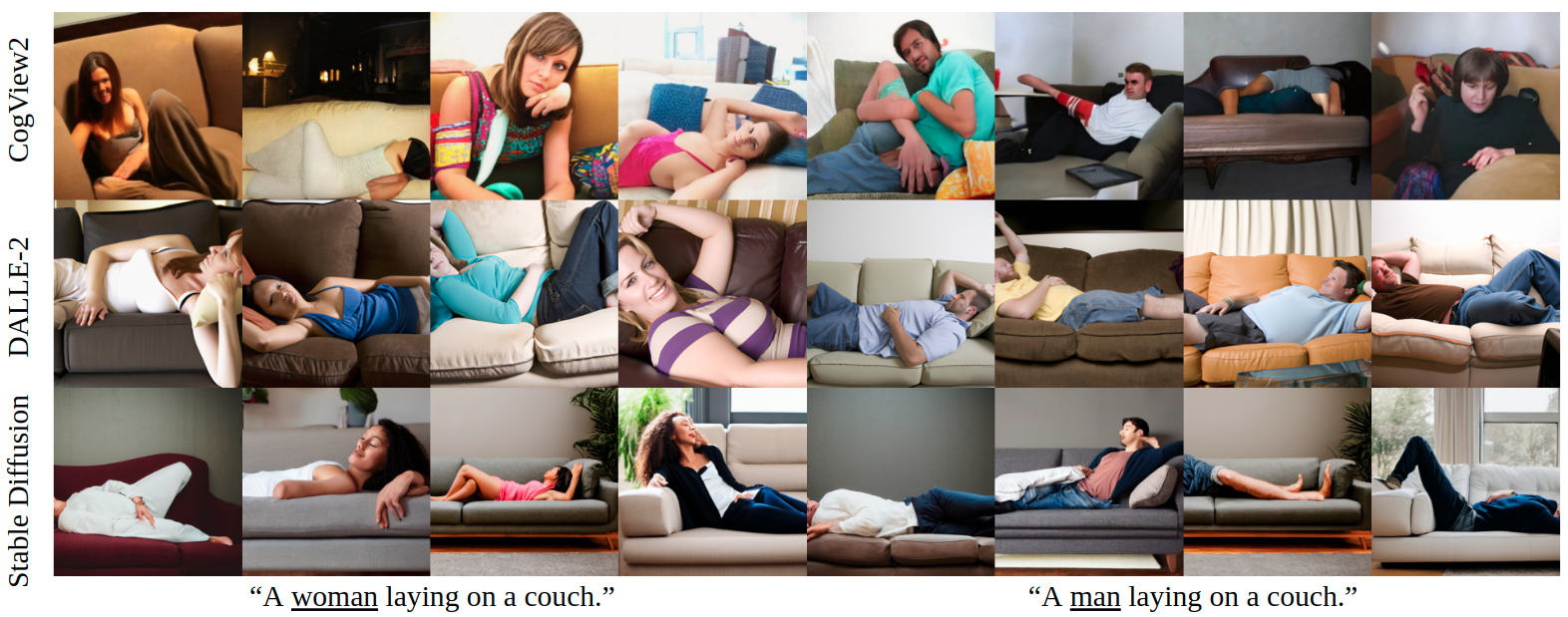}}
\caption{\label{fig-example-base-couch} Examples of gender presentation differences in the neutral setting. Images are generated from ``\emph{A woman/man laying on a couch.}''.}
\end{figure*}

\begin{figure*}[t]
\centering
\resizebox{2.0\columnwidth}{!}{\includegraphics{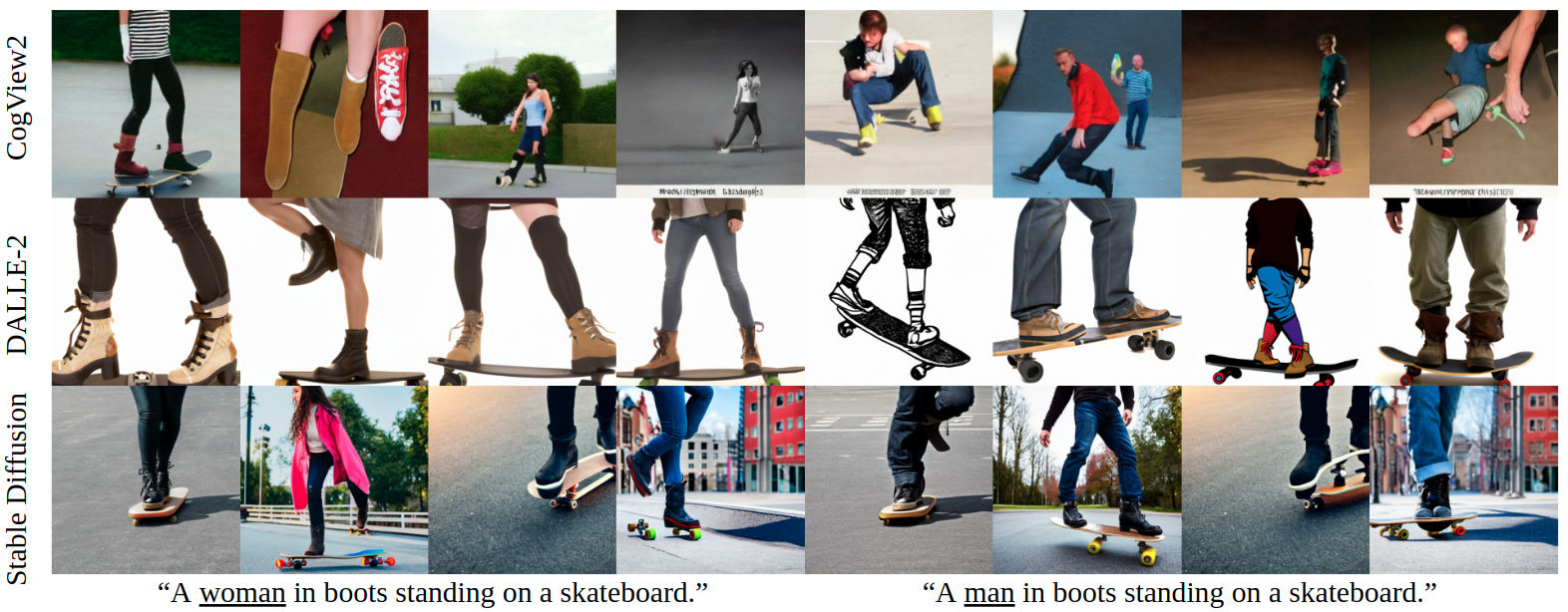}}
\caption{\label{fig-example-comb-boots} Examples of gender presentation differences in the explicit setting. Images are generated from ``\emph{A woman/man in boots standing on a skateboard.}''.}
\end{figure*}

\begin{figure*}[t]
\centering
\resizebox{2.0\columnwidth}{!}{\includegraphics{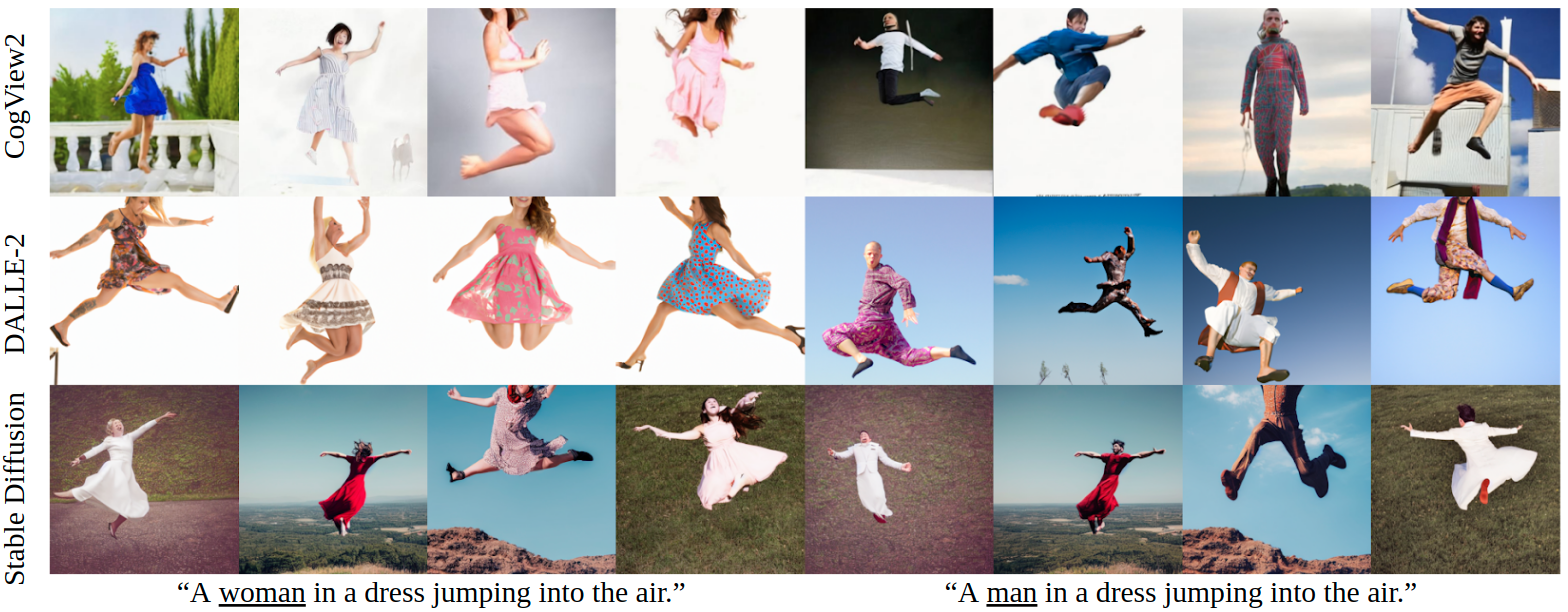}}
\caption{\label{fig-example-comb-dress} Examples of gender presentation differences in the explicit setting. Images are generated from ``\emph{A woman/man in a dress jumping into the air.}''.}
\end{figure*}

\end{document}